\documentclass[sigplan,10pt]{acmart}
\renewcommand\footnotetextcopyrightpermission[1]{}

\AtBeginDocument{%
  \providecommand\BibTeX{{%
    \normalfont B\kern-0.5em{\scshape i\kern-0.25em b}\kern-0.8em\TeX}}}

\usepackage{tikz, filecontents}
\usepackage{microtype,graphicx,xspace,booktabs,fancyvrb,multirow,color,soul,amsmath,etoolbox,arydshln}
\usepackage[inline]{enumitem}
\usepackage[export]{adjustbox}
\usepackage{comment}
\usepackage{hhline}
\usepackage{epsfig, subcaption}
\usepackage[rightcaption]{sidecap}
\usepackage[font=small, textfont={up}, skip=4pt]{caption}
\hypersetup{colorlinks=false, pdfborder={0 0 0}}

\providetoggle{editing_mode}
\settoggle{editing_mode}{true} 

\graphicspath{{./fig/}}
\DeclareGraphicsExtensions{.pdf,.png,.eps}
\newcommand{\sys}{\mbox{Canoe}\xspace}

\newcommand{\lng}{{logical neighbor}\xspace}

\definecolor{darkred}{rgb}{0.5, 0, 0} 
\definecolor{darkgreen}{rgb}{0, 0.5, 0} 
\definecolor{darkblue}{rgb}{0,0,0.5} 
\newcommand{\todotemplate}[3]{%
	\mbox{}
	\marginpar{%
		\colorbox{#2!80!black}{\textcolor{white}{#1}}%
		\vspace*{-20pt}
	}%
	\textcolor{#2}{{#3}}%
}

\newenvironment{tightitemize}%
 {\begin{list}{$\bullet$}{%
 		\setlength{\leftmargin}{10pt}
        \setlength{\itemsep}{0pt}%
        \setlength{\parsep}{0pt}%
        \setlength{\topsep}{0pt}%
        \setlength{\parskip}{0pt}%
        }%
 }%
{\end{list}}
\newcounter{tecounter}
 {\begin{list}{\arabic{tecounter}.}{%
 		\usecounter{tecounter}
 		\setlength{\leftmargin}{10pt}
        \setlength{\itemsep}{0pt}%
        \setlength{\parsep}{0pt}%
        \setlength{\topsep}{0pt}%
        \setlength{\parskip}{0pt}%
        }%
 }%
{\end{list}}%
 {\begin{list}{}{%
 		\setlength{\leftmargin}{10pt}
        \setlength{\itemsep}{0pt}%
        \setlength{\parsep}{0pt}%
        \setlength{\topsep}{0pt}%
        \setlength{\parskip}{0pt}%
        }%
 }%
{\end{list}}%

\iftoggle{editing_mode}{
\newcommand{\ada}[1]{\todotemplate{Ada}{cyan}{#1}}
\newcommand{\hd}[1]{\todotemplate{HD}{orange}{#1}}
\newcommand{\yc}[1]{\todotemplate{YC}{brown}{#1}}

\newcommand{\TODO}[1]{\textcolor{red}{#1}}
\newcommand{\todo}[1]{\textcolor{red}{#1}}
\newcommand{\fixme}[1]{\textcolor{red}{#1}}
\newcommand{\fixed}[1]{\textcolor{red}{\st{#1}}}
\newcommand{\discussion}[1]{\textcolor{orange}{#1}}
\newcommand{\camera}[1]{\textcolor{blue}{#1}}
}{
\newcommand{\ada}[1]{}
\newcommand{\hd}[1]{}
\newcommand{\yc}[1]{}
\newcommand{\TODO}[1]{}
\newcommand{\todo}[1]{}
\newcommand{\fixme}[1]{}
\newcommand{\fixed}[1]{}
\newcommand{\discussion}[1]{}
\newcommand{\camera}[1]{}
}

\usepackage{tikz}
\usepackage{xcolor}

\newcommand*\hollowcircled[1]{\tikz[baseline=(char.base)]{
            \node[shape=circle,draw,inner sep=1pt] (char) {\textcolor{black}{#1}};} }

\begin{document}
\title{\Large \bf Canoe : A System for Collaborative Learning for Neural Nets}
\author{
	Harshit Daga \qquad Yiwen Chen \qquad Aastha Agrawal \qquad Ada Gavrilovska\\Georgia Institute of Technology
}

\begin{abstract}


For highly distributed environments such as edge computing, collaborative learning approaches
  eschew the dependence on a global, shared model, in favor of 
 models tailored for each location. 
 Creating tailored models for individual learning contexts reduces the
 amount of data transfer, while collaboration among peers provides
 acceptable model performance.
Collaboration assumes, however, the availability of knowledge
transfer mechanisms, 
which are not trivial for deep learning models where knowledge isn't
easily attributed to precise model slices. 
We present
\sys~-- a framework that
facilitates knowledge transfer for neural networks.
\sys  
provides new system support for dynamically extracting  significant
parameters from a helper node's neural network, 
and uses this 
with a multi-model boosting-based approach to improve the predictive performance of the target node.
The evaluation of \sys with different PyTorch and TensorFlow neural
network 
models demonstrates that the knowledge transfer mechanism improves
the model's adaptiveness to changes up to {$3.5\times$} compared to
learning in isolation, while affording several magnitudes 
reduction in data movement costs
compared to federated learning. 


\end{abstract}

%
%
%
%
%
%
%
%

\settopmatter{printfolios=true}
\settopmatter{printacmref=false}
\maketitle
\pagestyle{plain}


\section{Introduction}
\label{section:introduction}

\vspace{-1.0ex}
\noindent The rise of sensors and connected devices leads to tremendous growth in the amount of data generated at the edges of the network.
 In order to manage, process and analyze the increasing amount of data,
 service providers
 have started relying on machine learning techniques to provide automation and to improve user experience, content discovery, curated search results or recommendations 
 \cite{ml_fb,ml_netflix,ml_personalization,ml_twitter,ml_sales_practice,ml_microsoft}.

  With data been generated in a highly distributed manner, 
  companies build machine learning (ML) models using systems for distributed learning such as Federated Learning (FL)~\cite{konevcny2015federated, arivazhagan2019federated}.
  Such solutions leverage computational resources closer to the sources, in geo-distributed data centers~\cite{hsieh2017gaia} or at many distributed server locations in the edges of the network~\cite{fl_edge_caching,federatededge,colla}, 
  to learn in place from localized data and summarize the changes as a model update. Model updates from all participants are aggregated in a centralized location 
  to improve a global model, which is then shared with all.
  FL reduces the data transfer cost compared to approaches which aggregate the raw data centrally, however the communication with a remote cloud is still required for regular exchange of model updates. 

In fact, 
by aiming to build a global, generic model, 
all of the above approaches rely on periodic communication among distributed locations, 
in order to transfer raw data or model updates. 
Highly distributed environments, particularly such as those observed across 
nodes in mobile edge computing (MEC)~\cite{etsi:mec}, have been shown to benefit from models tailored to the context of specific locations~\cite{att_edge_data,cellscope, traffic_dynamics_celluar,cellular_traffic_predictability,cartel,colla}. In addition, distinct locations may not exhibit any change in their localized inputs, thus not benefiting from model updates due to churn in data trends observed at other locations. Abandoning any cross-node coordination in learning in favor of isolated learning approaches, eliminates data transfer costs. However, if data trends shift across locations (i.e., in the event of concept drift), models learned in isolation suffer from loss of accuracy and need more time to adapt to change, compared to when relying on a global model. 
In response, distributed machine learning models for {\em collaborative learning} across edge nodes have been proposed, as a way to create robust tailored models.
Collaborative learning 
allows distributed nodes to adapt quickly to input shifts and retain model accuracy, 
with low data transfer costs~\cite{cartel,colla}.




A key assumption in realizing collaborative learning is the ability to transfer knowledge from one peer who has previously learned a given input class -- {\em a helper node} -- to another peer exhibiting a change in its inputs -- {\em a target node}.
For certain types of machine learning models, such as the online random forest (ORF) and online support vector machines (OSVM) used in~\cite{cartel}, realizing knowledge transfer is straightforward, as the association of input classes to model parameters is easily identifiable: to transfer knowledge about a given class from a helper to a target model, the corresponding subtrees in ORF or matrix rows in OSVM models need to be copied over across models. In practice, however, services are increasingly turning to deep learning techniques based on neural networks~\cite{apple:siri, salesforce:aieconomist, fb:deeptext, google:gboard, Bhatia2020AWSCS, uber, DeepFovea}, where knowledge about a particular input class cannot be trivially attributed to a slice of the model. This raises the question, {\em how to enable knowledge transfer across neural network (NN) models?}
Without the support to extract features from a helper node's model relevant for the knowledge needed at the target, and to then integrate them with the target node's model,  collaborative learning will not be feasible for deep learning techniques. 
This will limit the opportunity for the benefits of collaborative distributed learning
to be afforded to  many important application classes relying on deep learning.

In response, 
we present {\bf \sys } -- a framework that enables
{\em collaborative deep learning} by facilitating 
selective knowledge transfer across neural network models.
\sys
combines system support for dynamic creation of helper models from an existing neural network model, 
with 
a multi-model boosting-based approach to use this knowledge and update a target model.
The knowledge transfer in \sys is made possible through the use of mechanisms for (1) on-demand gleaning of {\em significant parameters} of a model, (2) use of {\em boosted helper models}, and (3) support for {\em managing helper model chains}. 
\sys is evaluated for several neural network models 
based on PyTorch and TensorFlow, and shows opportunities for significant improvements in model performance and data transfer costs, across both frameworks and with different models. 
The evaluation also demonstrates that \sys exposes new opportunities to exploit tradeoffs among model accuracy, the ability to quickly adapt to changes in a model's predictive performance, 
data transfer costs and learning overheads. Future policy engines can exploit such tradeoffs to further tailor the behavior of the distributed learning process.



\noindent In summary, the contributions of this paper include: 

\begin{tightitemize}
\item New mechanisms to facilitate knowledge transfer for deep learning models (\S\ref{section:mechanisms}); 
\item The design and implementation of \sys, a first framework that integrates support for collaborative deep learning~(\S\ref{section:runtime_implementation}); 
\item Evaluation with several models, ML frameworks, and workloads, which 
  demonstrates that \sys enables collaborative deep learning 
  across distributed nodes in a manner that allows models to adapt to change $3.5\times$ faster than isolated learning while giving a similar predictive performance compared to federated learning,  
  but with orders of magnitude lower overall data transfer costs (\S\ref{section:evaluation}). 
\end{tightitemize}

\section{Background and Motivation}
\label{section:background_motivation}
\begin{figure*}[ht!]
    \centering
    \includegraphics[width=\textwidth]{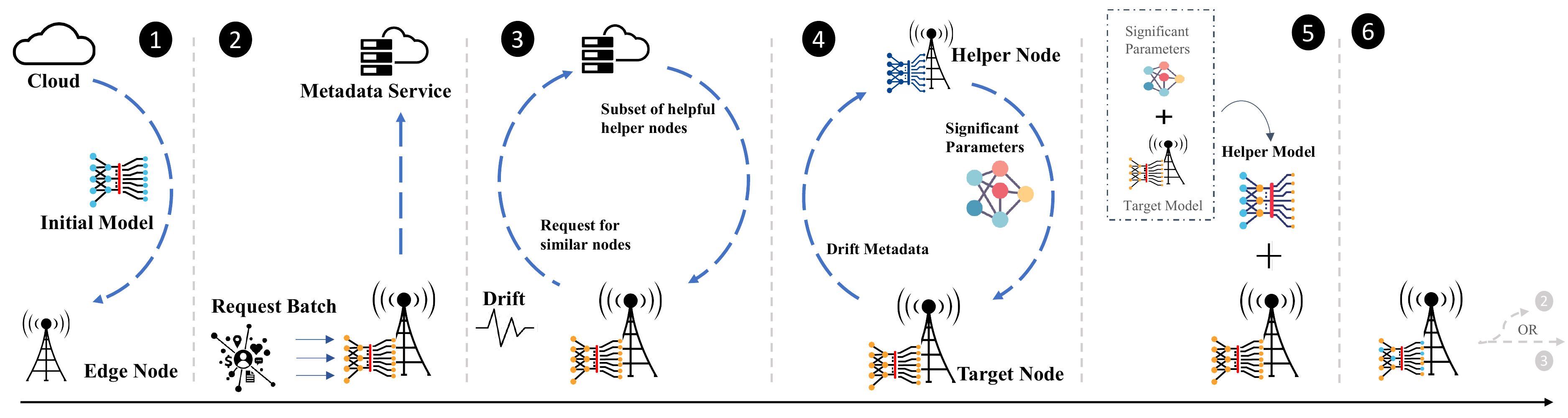}
    \caption{Collaborative learning in distributed edge computing.} 
    \label{fig:overview}
  \end{figure*}

  
\noindent{\bf Multi-access Edge Computing.}
MEC~\cite{etsi:mec,sabella_2016} provides IT services and compute capabilities at the edge of the network, in wireless and cellular access points such as gateways and cell towers,
in close proximity to the client devices. 
It offers infrastructure for execution of applications requiring low latency (e.g., AR/VR, automation~\cite{ar_offloading, ren_2019}), reduction in backhaul data demand (e.g., IoT 
analytics~\cite{dnn_offloading_edge,satyanarayanan2015edge}, content delivery~\cite{Fajardo_2015, zhang_2018}), and geo-localized data handling (e.g., for GDPR~\cite{gdpr}).
One challenge MEC exposes is that of creating an extremely dispersed distributed infrastructure, which differs from datacenter-based distributed environments in its scale and in the fine-grained geo-localization of the inputs observed at each node~\cite{att_edge_data}. 
%
For instance, COLLA~\cite{colla}, a collaborative system to predict user behavior patterns, highlights that
a shared global model might not achieve the desired performance for all the users. Thus it trains smaller more tailored models
at individual edge nodes and captures users with similar behavior patterns. 
Cellscope~\cite{cellscope} also demonstrated the ineffectiveness of using a shared global model for managing the base stations in mobile network;  
similar observations are confirmed in other research~\cite{traffic_dynamics_celluar,cellular_traffic_predictability}.


\noindent{\bf Collaborative Distributed Learning.}
Contrary to federated learning, where a global shared model is created and distributed, collaborative distributed learning is relevant for contexts such as MEC, where information is localized and nodes require only a portion of the global model~\cite{cartel,colla}. This helps in reducing the overall data transfer, provides smaller model size and reduces (re)training time. 

Figure~\ref{fig:overview} illustrates a typical workflow in a collaborative system. Edge nodes use their resident model to perform inference, following which the
model is re-trained locally.
Upon 
shift in the data distribution, the model performance at the edge node deteriorates.
%
In such a scenario, the node relies on the collaborative system to find a helper node, and to leverage the helper's model (knowledge) to boost its own model. 
In order to find a helper node in a timely manner,
a common metadata service (MdS) aggregates information from each node in the system.
This metadata is information that helps the system differentiate the nodes, such as node hardware, network configuration, workload characteristics such as estimates of class priors, overall model performance, etc.
Each edge node uses drift detection algorithms~\cite{dd_threshold,kifer_window_dd,adaptive_dd} on the local metadata information to decide when a model at the target edge node needs to be updated, and to query the MdS for candidates for a possible helper. The MdS uses the metadata it has aggregated to determine helper node candidates using algorithms such as~\cite{KL_divergence, hellinger_distance, chi_squared, JS_Divergence}. 

In order to update the target node's model and improve its predictive performance, a collaborative system uses 
knowledge transfer~\cite{cartel} or distillation~\cite{colla}.
Knowledge distillation~\cite{distillation} uses a teacher model to train a student model. It is a process of transferring knowledge from a large model (teacher) to a smaller model (student) to obtain similar accuracy as the teacher model. Knowledge transfer is a mechanism to transfer the knowledge regarding several (typically not all) classes from one machine learning model to another. In this paper we use knowledge transfer mechanism to collaborate among edge nodes which use the same neural network. We chose knowledge transfer over distillation because in distillation the teacher model is assumed to be a perfect model,
whereas, we already point out, a perfect global model may not be needed for MEC, and it may be too costly to maintain it at large scale in a highly distributed settings. 

\noindent{\bf Neural Networks for MEC.}
When compared with other machine learning models, Neural Networks 
have shown superior performance for many emerging intelligent applications. Convolutional Neural Networks (CNN)~\cite{lecun1995convolutional} have shown best performance for image and video analytics, while Recurrent Neural Networks (RNN)~\cite{rnn} are extensively used for voice search~\cite{rnn_for_voice} such as Apple's Siri and Google's voice search. A growing number of applications
based on real-time analytics of sensor data, automation, AR/VR, etc.,~\cite{nn_video_service_realtime,dnn_offloading_edge,live_video_analytics, smart_parking,music_cognition_edge, edge_vr_dof, edge_ar}, 
have started to deploy deep learning services at the edge of the network. The real-time nature of applications makes the inference latency-critical. The large size and distribution of the training data that may be gathered across the user contexts of these applications, make such applications bandwidth intensive.
One potential solution is to use federated learning for such applications. However, when working with neural nets even small amounts of raw data can result in large amounts of information being communicated to the aggregating node~\cite{fl_large_scale}. The combination of these factors makes support for
collaborative learning for neural network an important distributed learning strategy. This, in turn, requires support for knowledge transfer. 

\noindent{\bf Knowledge Transfer in Neural Networks.} 
%
%
Knowledge transfer for neural networks is a hard problem that is receiving a lot of attention from the ML community~\cite{zhou2018revisiting, bau2018identifying,how_transferable_are_features_dnn}.
Even when considering
the same type of neural networks,  given the many connected layers, information required to classify a class is spread across each layer, requiring specialized algorithms to extract it. If the data observed at target node is very similar to the one observed by the helper, fine-tuning~\cite{cnn_fulltraining_fine_tuning} is one methodology 
that can be used, where the helper model design and parameters are fully or partially replicated on the target model. However, as workload characteristics
can vary across different nodes, using this technique can further reduce the model performance, as we show in \S\ref{subsection:applying_significant_param}. 
One alternative is to improve and speed up the learning of the target node model for the new task using techniques such as ~\cite{actor_mimic, ktm-drl}. However, such techniques assume the helper model to be an expert teacher model, which in our case might not be always feasible. 
A different approach is to apply techniques that extract the portion of the model that most contribute to its performance for a given input. 
\cite{zhou2018revisiting} suggests removing one neuron at a time to determine if it has impact on the model performance. \cite{yu2018distilling} analyzes the activation, gradients and visualization patterns to determine a critical path.
However, as the machine learning models are continuously (re)trained on the incoming data stream at each edge node, knowledge transfer needs to be performed dynamically, 
and over fresh models, requiring time and resource-efficient techniques. 

\section{Goal and Challenges}
\label{section:challenges}

\noindent{\bf Goal. }
The goal of \sys is to provide a collaborative framework for distributed deep learning 
which will enable 
fast knowledge transfer across the neural network models of the peer nodes.

\noindent{\bf System model. }
The 
previous section illustrates that 
in a collaborative system each edge node creates a customized model and requests for help when there is drift in the model performance. 
The concept of providing help is the ability to perform knowledge transfer from a helper node to a target node.
%
We assume  that helper nodes can be identified from the model and node metadata, using techniques such as~\cite{KL_divergence, hellinger_distance, chi_squared, JS_Divergence}, as done in~\cite{cartel}. 
The specific contributions made by \sys are new techniques which enable knowledge transfer to be performed across neural network models among distributed nodes 
in a manner that boosts the overall accuracy at the target node while maintaining low data transfer costs.  


\noindent{\bf Challenges. }
The fundamental issue in performing knowledge transfer across collaborative peer nodes lies in the fact that the helper node's knowledge is neither all necessary at the target node, nor is it strictly a superset of the target node's knowledge. Simply using the helper node's model introduces either unnecessary data movement (for portion of the knowledge not relevant for the inputs at the target), 
or worse, degrades the target node's knowledge (by displacing localized knowledge that was already acquired).
This is in contrast to centralized or federated learning frameworks, where a centralized helper node is assumed to have globally relevant, generic model. 
Instead, for collaborative learning only a relevant portion of the helper node's model needs to be transferred, such that the target model can achieve the needed boost in its accuracy, while also keeping the data transfer costs low. Addressing this issue for  neural network  models introduces the following main challenges:

{\bf C1:} {\em How to extract a relevant portion of a neural network model?}
    A neural network model consists of multiple hidden layers. It takes an input, passes it through multiple layers of hidden neurons (mini-functions with unique coefficients that must be learned), and outputs a prediction representing the combined input of all the neurons. Thus, various parameters spread across different layers help the model in providing a good predictive performance. The challenge is to determine a methodology that could help in selecting the relevant parameters from all the layers, and to do so with low overheads and with low resulting data transfer demands.

{\bf C2:} {\em How to utilize the knowledge at the target node?}
The target node's neural network similarly consists of hidden layers whose parameter values reflect existing locally acquired knowledge. The model parameters then need to be carefully updated with the knowledge obtained from the helper, in order to improve the target model without displacing local knowledge.
The challenge is to determine a methodology for updating the target node's model parameters with the helper's knowledge quickly and effectively, without degrading its performance on previously known classes. 

\section{Overview of \sys}
\label{section:clue}

\noindent{\bf Approach. }
To address these challenges, \sys adopts
an approach that combines support to {\em dynamically extract significant model parameters}, with use of {\em multi-model boosting}. Specifically, \sys relies on helper nodes, but only extracts the portion of their knowledge relevant for the target node.
It then uses boosting to create custom helper models for the target, and then uses those helper models, in conjunction with the target's existing knowledge, to create a more effective combined model. 

By using a 
multi-model approach, \sys makes it possible
to decouple the updates that  need to be performed to improve the target model's prediction accuracy from the specific parameters extracted from the helper node's original model.
By using a
simple boosting technique to create the helper models, \sys makes it possible to achieve a resulting helper model that benefits from the helper for recognizing new classes, while retaining its knowledge quality for pre-existing, familiar classes.


\noindent{\bf Mechanisms. }
Underpinning 
\sys 
are three key mechanisms designed for {\bf selecting significant parameters} from the helper node's model, {\bf creating a boosted helper model} and {\bf managing model chains} on the target node. 

To solve the challenge~{\em C1} of selectively picking the parameters from each layer of the neural network, the framework uses a mechanism for significant parameter selection, \hollowcircled{4} in \autoref{fig:overview}.  
The task of selecting significant parameters of a DNN model can be rephrased into the following question: given a DNN $f$ and input image $x$, what is the part of $f(x)$  that made the DNN reach the particular decision of classifying the input as class $y$. In other words, which coefficients or weights at each layer in the model contributed towards the classification of $x$ to an output $y$?

On the target node, \sys creates helper models using the significant parameters from the original model~\hollowcircled{5}.
A helper model with poor performance for some classes can degrade the overall 
model performance, effectively displacing knowledge that already existed as part of the target model.
To prevent this, a helper model should be created by
using 
the knowledge obtained from the helper node with the existing knowledge on the target node.
The resulting helper model is then used in conjunction with target model.

A target node may exhibit multiple instance of input drift over time. This will require
new knowledge transfers, 
and can result in a chain of helper models. 
To prevent helper models from creating a resource bloat on the target,
they are ephemeral, and are discarded when no longer needed. 
The target node continues to learn 
during the boosting period, so as to make it possible to remove its dependence on the helper model. Once all helper models are discarded, the target continues to operate only with its, now updated, target model \hollowcircled{6}. 
The use of a boosted helper model 
and the techniques for managing helper model chains, solve the challenge~{\em C2}.

\section{Design of \sys}
\label{section:mechanisms}
Next, we describe in greater detail the design of the three mechanisms \sys uses to realize knowledge transfer from a helper to a target node for neural network models. 

\subsection{Significant Parameters Selection}
\label{subsection:significant_param_selection}

To address the challenge {\em C1}, \sys provides support for extracting a relevant portion of a helper node's model in a practical manner via a mechanism for selecting the model's significant parameters.
%
A neural network model consists of multiple hidden layers of neurons, each with their activation function, interconnected with weighted synapses. During the learning process each neuron takes a group of weighted input, applies an activation function and returns the output.
These inputs can be features from the dataset or output of a previous layer's neuron.
The weights are adjusted using backpropagation~\cite{backpropagation} ("backprop"), and the process is repeated 
until the network error drops below a 
threshold.
Thus, for a given input, a combination of coefficients across multiple layers plays a role in determining the outcome.
Our goal is to select from each layer of a model the significant parameters that are relevant to the requested classes.

A trivial design choice could be to send the entire model from a helper to a target node while avoiding any overhead of selecting a relevant portion of the helper model. However, this is not a good design choice as NN models are large, and can include millions of parameters as shown in Table~\ref{table:per_request_data_transfer}.
Furthermore, when considering collaborative learning, only certain parameters relevant for the missing classes are required by the target node. Thus, by pre-processing the helper's node original model to identify and send only the significant parameters for the relevant classes,
the data transfer cost are reduced.
\noindent\textbf{Parameter Sensitivity. }
To select the significant parameters we first need to 
determine the relation and influence of each parameter on the output value, i.e., the  \textit{parameter sensitivity}. 
A neural network consists of a large number of parameters. A change in the value of these parameters has a direct impact on the predictive performance of the model. Our goal is to first understand the sensitivity and impact of each parameter on the output value, and then to determine a subset of the most significant parameters. 
We use the backprop technique to determine the sensitivity by finding the impact that a small change in an input neuron ({\em i})
has on an output neuron  ({\em o}).
If drastic change occurs, {\em i} is considered to be significant for producing the current activation value of {\em o}. 

Let $f(W, x) = p$ denote the neural network model where $W$, $x$ and $p$ are set of all the model parameters, inputs and prediction, respectively. For a multi-class classification task the output of the neural network is an array of size $C$ where each element represents the confidence of the input being class $C_i$. We determine the sensitivity value of a parameter $W_j$ towards the class output value $C_i$ as
\begin{equation}
    sen(W_j, C_i) = \frac{\partial C_i}{\partial W_j} = \lim\limits_{\Delta W_{j} \rightarrow 0} \frac{C_i(W_j+\Delta W_{j}) - C_i(W_j)}{\Delta W_{j}}
  \end{equation}
  
$sen(W_j, C_i) > 0$ means increasing the value of parameter $W_j$, increases the output value for class $C_i$ which implies the model is more confident that the input belongs to class $C_i$. On the contrary, $sen(W_j, C_i) < 0$ means increasing 
$W_j$ makes the model more confident about an input not being class $C_i$.
Thus, $|sen(W_j, C_i)|$ is defined as the measure of sensitivity of the parameter. The larger magnitude indicates higher significance of the parameter, as a small change in the parameter value could change the output. 

Calculating the sensitivity value of each parameter for all the classes is computational expensive. 
Thus for a class the sensitivity of each parameter is calculated using the backprop technique similar to the one used in gradient descent~\cite{sgd_backprop} where it uses the partial derivative to determine the impact of change of the parameter on the class output value.

\noindent\textbf{Determining sensitivity for a changing model. } Over 
time the model at the edge node changes. As a result significant values cannot be calculated in the initial phase and used at a later time when a request from the target node is received.
As the model is continuously retrained there are few ways to evaluate these values~\cite{zhou2018revisiting,sa_perturbation,sa_weight_perturbations,natural_gradient}. We can either pre-calculate the sensitivity after every batch in a {\em continuous} manner, and then do a quick look up when a request from a target node is received, or dynamically compute these values upon request. To evaluate the contribution of parameters {\em on-demand}, we can either add Gaussian noise at different layers~\cite{progessive_nn} or have a data reservoir and use the backprop technique to measure 
how their change contributes to the model output ($o$).

The first method is based on a {\em continuous} update of a sensitivity map for all model parameters and classes. Since the gradient is linear we combine the neural net output of all the data points in a batch to compute the parameters sensitivity value for all the classes. For the subsequent batches the values are recalculated and are added with the previously stored values. After every batch, this calculation of parameters adds a compute overhead of up to $0.7\times$ the time taken to process a batch. Additionally, since the sensitivity value of a parameter is different from its actual value in the model, this approach requires an addition memory of $O(CW)$ at each edge node. With  large NN model sizes and resource constrained edge nodes, this method might not be a viable option. However, the pre-calculated values help in quickly responding to the request when a request for a portion of the model is made. 

A second approach is one where significant parameters are calculated 
{\em on-demand}, when a 
request is received. 
To achieve this we keep a smaller local database (data reservoir) of the last $B$ request batches. When requested, the sensitivity of the parameters is calculated against the stored batches in a similar way as before. The effectiveness and overheads of this method are dependent on 
the exact setting of $B$. In general, the choice depends on the workload dynamics and model. For our dataset and models we experimentally found that $B= 1$ worked well as each request batch consists of sufficient data points.

The above two approaches showcase a tradeoff between the amount of memory required to calculate the sensitivity parameters and response time. \S\ref{eval:overheads} discusses the overhead involved in each of the approaches.
\sys provides a \textit{pluggable parameter} selection component which enables the user to use any technique. 

\noindent{\bf Significant Parameters. }
Once the helper node determines the sensitivity of the parameters for the requested classes, 
the system selects the top $Z$ percent of the sensitive parameter (both positive and negative). For a multi-class request we average the sensitivity value for all the classes before selecting the top $Z$ parameters. After selecting the top  parameters, \sys selects the corresponding weights of these parameters from the model. Since these parameters are the most relevant towards the classification of a class we refer to them as {\em significant parameters}. The significant parameters are compressed, which reduces the data size by up to $3$-$5\times$, and are sent to the target node. The exact setting of $Z$ depends on the model as well as the tradeoff between the data transfer versus knowledge transfer to improve the target node model. For our dataset and NN models we experimentally found that $Z$ can vary in the range $20\%$ to  $50\%$. \S\ref{eval:tuning} illustrates the impact of different values of $Z$ and we leave the automatic tuning of $Z$ as a topic for future investigation.

\subsection{Helper Model}
\label{subsection:applying_significant_param}

\begin{figure}[t!]
\centering
 
 \begin{subfigure}[b]{0.95\columnwidth}
    \includegraphics[width=0.95\columnwidth]{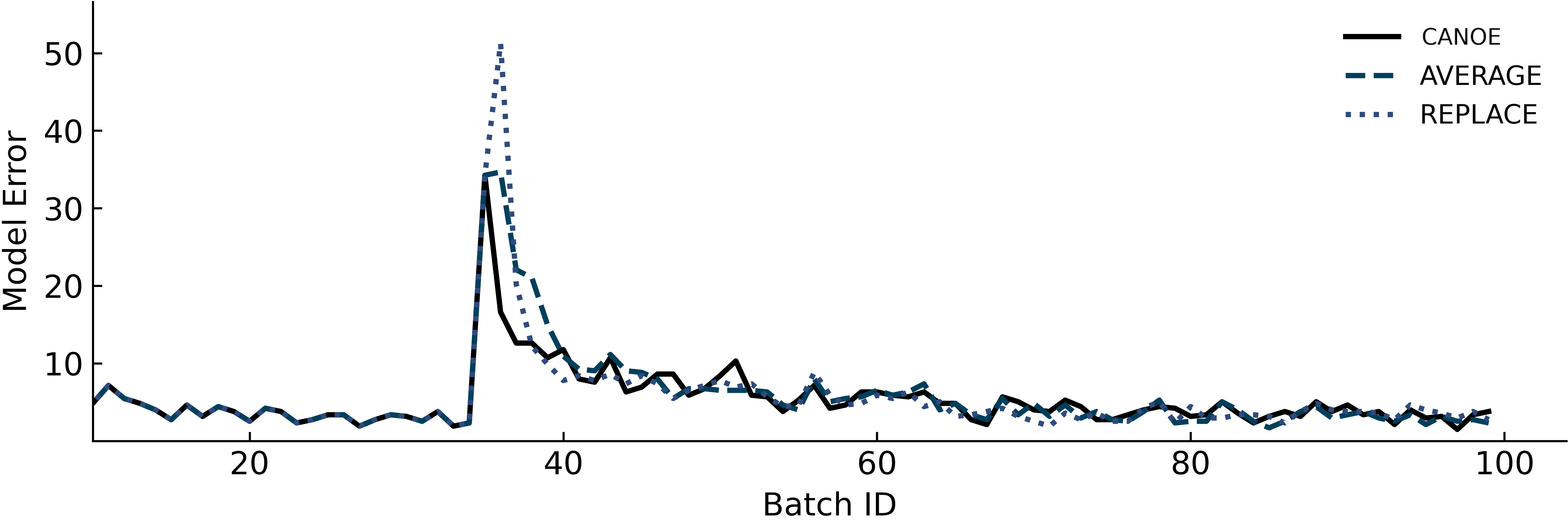}
    \caption{Non overlapping workload}
    \label{fig:non_overlapping_avg_replace_sp_comparision}
 \end{subfigure}
 
 \begin{subfigure}[b]{0.95\columnwidth}
    \includegraphics[width=0.95\columnwidth]{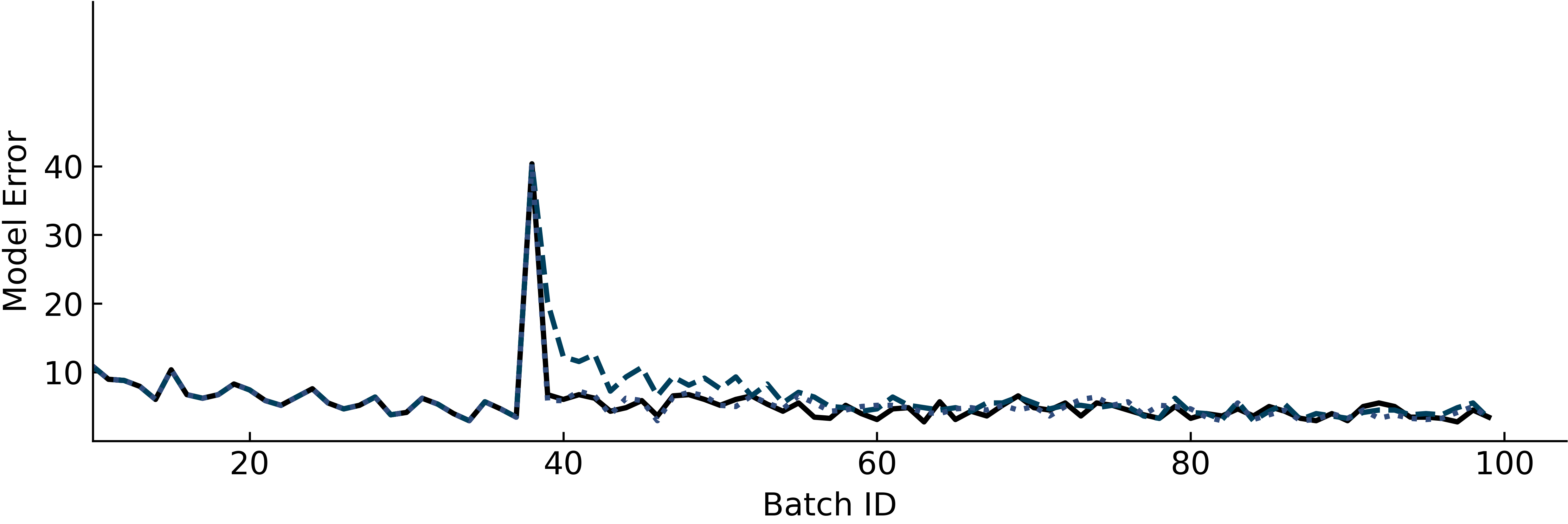}
    \caption{Overlapping workload}
    \label{fig:overlapping_avg_replace_sp_comparision}
 \end{subfigure}
 
\caption{Multi-model approach compared with averaging or replacing the existing model parameters: \sys performed well and was consistent in both the scenarios; replacing the parameters worked well with overlapping classes and average performed better when the classes at the two nodes were non-overlapping.}
\label{fig:avg_replace_sp_comparision}
\end{figure}

 To address the challenge {\em C2}, \sys provides support for creating and managing helper models used for boosting the model at the target node.
  
 After receiving the significant parameters from the helper node(s), the target node needs a mechanism to apply these updates to improve the predictive performance of its model. Let $W^t$ and $W^s$ represents the target node and significant parameters, respectively. A simple approach would be to apply the received knowledge directly on the existing target model. Different from replacing the significant parameters is to create the model by averaging the weights of target model with the received significant parameters. This technique would be similar to Federated average algorithm~\cite{mcmahan2017communication}.

 The outcome of replacing or averaging the existing parameters is dependent on the workload distribution.
 Figure~\ref{fig:avg_replace_sp_comparision} illustrates the model error rate observed over the 100 request batches from a subset of the classes in the EMNIST~\cite{emnist} dataset at the target node 
 using the MobileNet~\cite{mobilenet} convolution neural nets for image classification. The workload distribution observed at different nodes in the system can vary.
 It can be 
 overlapping where the workload observed at the target 
 is a subset of the workload observed at another node, or non-overlapping where no two classes observed at the nodes are the same.
 To illustrate the impact of replacing or averaging the existing parameters we use two workloads -- overlapping and non-overlapping. At batch id 35 a new class(es) is introduced at the target node. When the target node workload distribution is a subset of the workload observed at the helper node (overlapping), replacing the parameters would provide more benefits. However, when the workloads differ (non-overlapping), averaging the parameters would help more when compared to parameter replacement. Figure~\ref{fig:non_overlapping_avg_replace_sp_comparision} illustrates that for a non-overlapping workload replacing the parameters further increases the model error rate since it replaces parameters for classes already learned by the model. When the target and helper model observe a similar workload, Figure~\ref{fig:overlapping_avg_replace_sp_comparision} shows that replacing the parameters helps, whereas averaging the parameters would result in the target model taking more time to adapt to the new changes. 

\noindent\textbf{Multi-model approach. }
To decouple the way the significant parameters are used to update the target node's model from the dependence on how the workload is distributed across the target and helper node(s) in the system, we use the significant parameters to create a helper model. 

The existing model at the target node is well trained to predict the already observed classes, while the significant parameters received from the helper 
can assist in improving the prediction of the newly observed classes. As neither the target model nor the significant parameters can solely make accurate prediction, \sys uses a multi-model approach where a helper model is created 
using a boosting technique~\cite{dai2007boosting, adaboost}. Boosting
converts or combines weak learners to form a strong the helper model.
We use the original target model with the goal of maintaining the accuracy of the model for the existing classes. We use the helper model to provide a boost in accuracy for new classes which were already known at the helper node. 
The final output is a resultant of combining the output from both the models. 

\begin{figure}[th!]
    \centering
    \includegraphics[width=0.95\columnwidth]{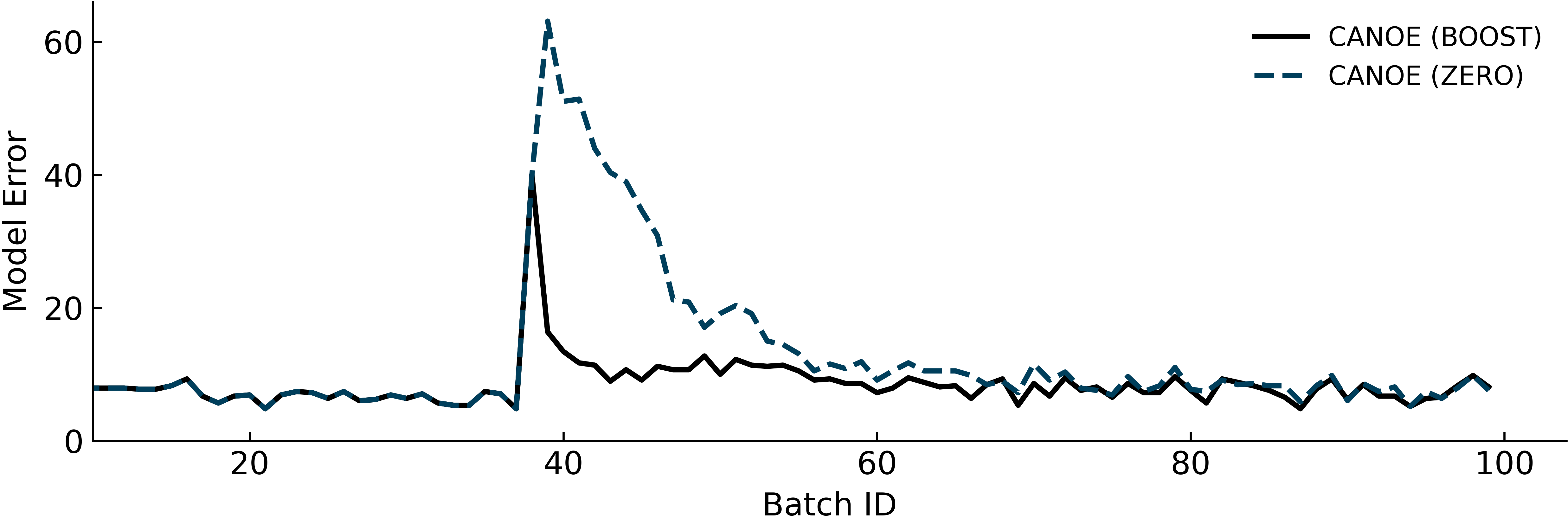}
    \caption{Boosted model error rate for two types of helper models. Helper model created with boosting from target model performs better than the one created only using the significant parameters.}
    \label{fig:boosting_vs_no_boosting_comparision}
\end{figure}

\noindent\textbf{Helper model creation. } As the target node receives only a subset of the parameters from the helper node(s), we consider two ways to create a helper model. In the first method, shown as \sys (ZERO) in Figure~\ref{fig:boosting_vs_no_boosting_comparision}, the helper model is created by using the significant parameters, and the remaining parameters of the model are set to $0$. Using $W^{h}$ to denote the parameters of the helper model, we formulate this as
\begin{equation}
    \begin{aligned}
        W^{h}_i \leftarrow
            \begin{cases}
                W_i^s,\quad & \textmd{if $W_i^s$ is significant}  \\ 
                0,   & \textmd{if $W_i^s$ is not significant}
            \end{cases}
    \end{aligned}
\end{equation}
Neural network consists of interconnected neurons where each connection is associated with a weight which helps in determining the output. By having some weights in the layer as $0$ effectively reduces the magnitude of the output of that layer, eventually impacting the overall output. In the worst case, if the parameters in one of the layers are all set to $0$, all intermediate values and final class output scores would be the same for any input. In such a case, no useful information could be derived from the helper model output. Therefore, we consider a second method, which creates a helper model by combining the significant parameters from the helper node with the existing parameters from the target node. The resulting {\em boosted
helper model} is represented as shown in Equation~\ref{eq:helper_model_boosting}, and illustrated as \sys (BOOST) in  Figure~\ref{fig:boosting_vs_no_boosting_comparision}. 
\begin{equation}
\label{eq:helper_model_boosting}
    \begin{aligned}
        W^{h}_i \leftarrow
            \begin{cases}
                W_i^s,\quad & \textmd{if $W_i^s$ is significant}  \\ 
                W_i^t,   & \textmd{if $W_i^s$ is not significant}
            \end{cases}
    \end{aligned}
\end{equation}
\noindent\textbf{Boosted model output. } After creating the helper model
we update the prediction function to take into account the classification outputs from the helper model. At this stage we have a target model, good for the existing classes, and a helper model, 
good for the new ones. 
We use average of both the models to compute the final output, though other functions can be plugged in. 

Figure~\ref{fig:boosting_vs_no_boosting_comparision} compares the two approaches of helper model creation, and shows that a helper model created using the target model is more effective when compared to the one created using no prior information: \sys (BOOST) experiences a lower spike in the model error rate when a previously known class is observed at the target, and it more quickly adapts 
and converges to its original accuracy. 

\noindent{\bf Managing helper model chains.}
\label{subsubsection:model_chaining}
The use of helper models improves the overall performance of the model, however, it creates the problem of repeatedly generating new helper models, on-demand, whenever a concept drift is detected. Maintaining an ever lasting 
chain of helper models 
would add more strain on the resources at the edge node. To alleviate this pressure we take two steps. First, since the goal of the helper model is to provide assistance to the target model, we freeze the helper model by not retraining it, 
while we continue to (re)train the target model. Second, we discard the helper model(s) as the 
performance of the model stabilizes for the class(es) for which the helper model was created.

A straightforward way would be to discard the model after a user-defined number of batches. Discarding a helper model prematurely could result in a second smaller drift, while discarding it too late would result in additional unnecessary computation and resource usage. To avoid this uncertainty \sys uses a dynamic approach where after every request batch the performance of the target model is tracked separately and is compared to the pre-drift performance. When the target model reaches the earlier performance level the helper model is discarded. In presence of chain of helper models \sys provides an additional check which compares the predictive performance of the new classes for past $L$ batches 
to determine which model to discard. The precise value of $L$ depends on the number of samples observed, models and dataset. For our experiments the value was experimentally determined and we leave the online tuning of $L$ as a topic for future work.
The precise discard policy 
is configurable. 

\section{\sys Runtime}
\label{section:runtime_implementation}

We 
describe the interaction among the different components in \sys, and discuss  the requirements for its implementation
in the context of existing 
ML frameworks.



\begin{figure}[t!]
    \centering
    \includegraphics[width=\columnwidth]{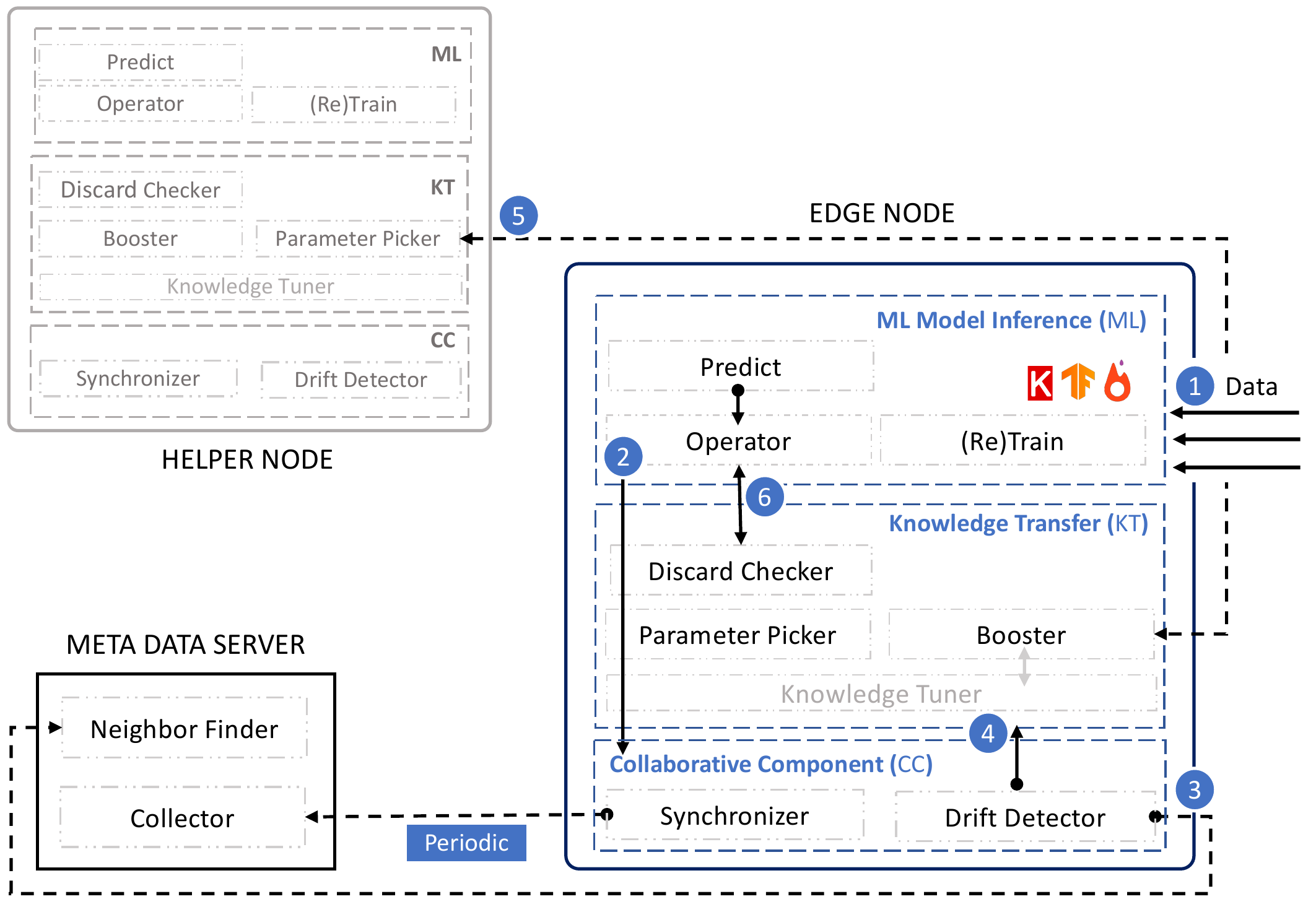}
    \caption{\sys System Components.}
    \label{fig:sys_components}
\end{figure}

\noindent{\bf \sys Components.} The runtime is divided into three main blocks -- ML Model Interface (ML), Knowledge Transfer (KT), and Collaborative Component (CC),  
shown in  Figure~\ref{fig:sys_components}.

The {\bf CC} component controls how a node collaborates with other nodes in the distributed system. It encapsulates the functionality needed to discover a helper node and to determine when to reach out for help. 
{\em Synchronizer} 
is responsible for exchange of 
metadata 
that provides the necessary state for helper node discovery. 
{\em Drift detector} determines when interactions with a helper node get triggered,  ideally, 
when the local model degrades in performance, 
based on a {\em drift detection} algorithm, as in~\cite{cartel}.
In~\cite{colla},
synchronization 
is performed periodically, and is always coupled with a receiving help via model distillation. 


The {\bf ML} component provides the interface to the machine learning framework (e.g., PyTorch or TensorFlow). It uses the \textit{predict} interface to perform the prediction on the incoming request data batch. The output of the prediction is passed to an \textit{operator} component. During normal operation, the prediction result is used directly. In the presence of helper model(s), during periods of knowledge transfer, the {\em operator} averages the outcome of all the model(s). 
The cycle is completed with training the resident target model with the new batch of data using the \textit{(re)train} interface. 


The {\bf KT} component facilitates the knowledge transfer mechanisms. At each node, {\em parameter picker} supports the parameter selection mechanism, and is invoked asynchronously, when a node receives a request for help for a specific class(es).
The process is performed by either the {\em continuous} or the {\em on-demand} mechanism. 
The resulting parameters are sent to the target node where {\em booster} creates  a new  helper model.
When 
helper model(s) are created, the \textit{operator} function is adjusted to perform averaging, and a \textit{discard checker} monitors the convergence process to determine when helpers are to be discarded, dynamically or statically after predefined number of batches, according to a configurable policy.


\sys exposes APIs to configure the system parameters controlling the knowledge transfer process in terms of 
the percentage of significant parameter transferred from the helper, the number of request batches to be locally stored for {\em on-demand} parameter selection,
the discard policy, etc. In this manner, \sys exposes new {\em tuning knobs} that control the knowledge transfer process in terms of its effectiveness in improving the target model and the overheads it introduces, which can be used for new types of resource orchestrators for distributed learning. 



\noindent{\bf Implementation.} We implemented a prototype of \sys
for the PyTorch and TensorFlow 
ML frameworks, using Flask to  provide  support for communication among different nodes.

The use of the different frameworks exposes the following requirement for implementing \sys. 
The significant parameter selection can be done in different ways, however using the dynamic selection approach would require the ML framework to provide corresponding APIs for calculating the gradients on the model. 
The framework should also provide the capability to dynamically create, use and discard helper model(s), in conjunction with the existing model. 
Given the available APIs in PyTorch, our implementation of \sys supports all the mechanisms described in \S\ref{section:mechanisms}. 
TensorFlow creates a static computation graph before a model is run and any interaction with the model is tightly integrated with the session interfaces and
state  placeholders.
%
Due to these limitations,  TensorFlow cannot support {\em on-demand} significant parameter selection, hence, this implementation of \sys uses 
the {\em continuous}
significant parameter mechanism. Additionally, since TensorFlow does not permit models to be dynamically created, we use a pool of helper models from the initial state of the experiment.
The resulting TensorFlow implementation of \sys introduces runtime memory overheads which may be prohibitive for practical settings. We note that the limitations of the current TensorFlow APIs may be addressed in future versions of the system, and we still compare the two implementations to illustrate the generality of \sys for realizing knowledge transfer across models developed with different ML frameworks (\S\ref{eval:frameworks}).  

\begin{figure*}[t!]
\centering
 \begin{subfigure}[b]{0.33\linewidth}
    \includegraphics[width=\linewidth]{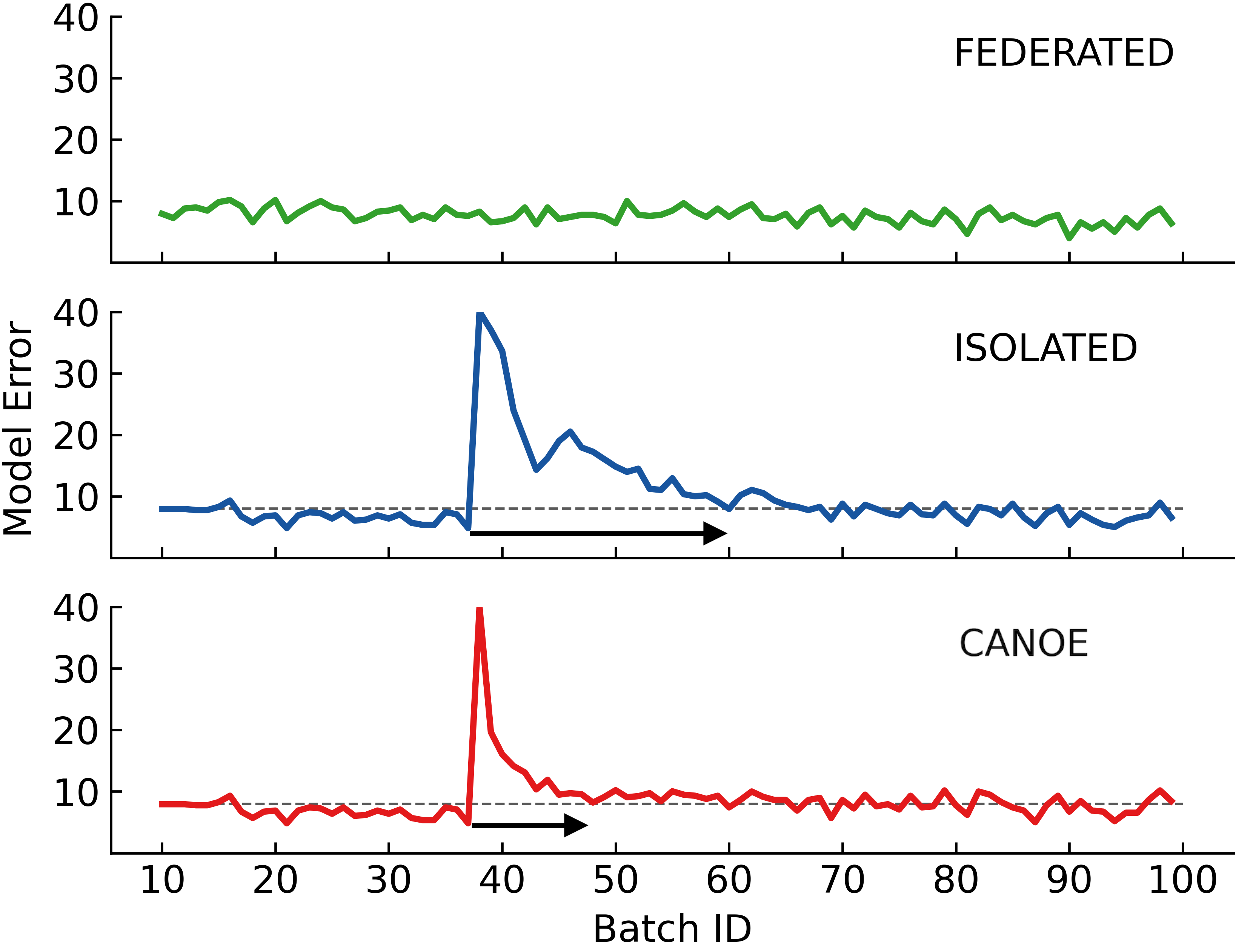}
    \caption{MobileNet}
    \label{fig:mobilenet_overall_intro}
 \end{subfigure}
 \begin{subfigure}[b]{0.33\linewidth}
    \includegraphics[width=\linewidth]{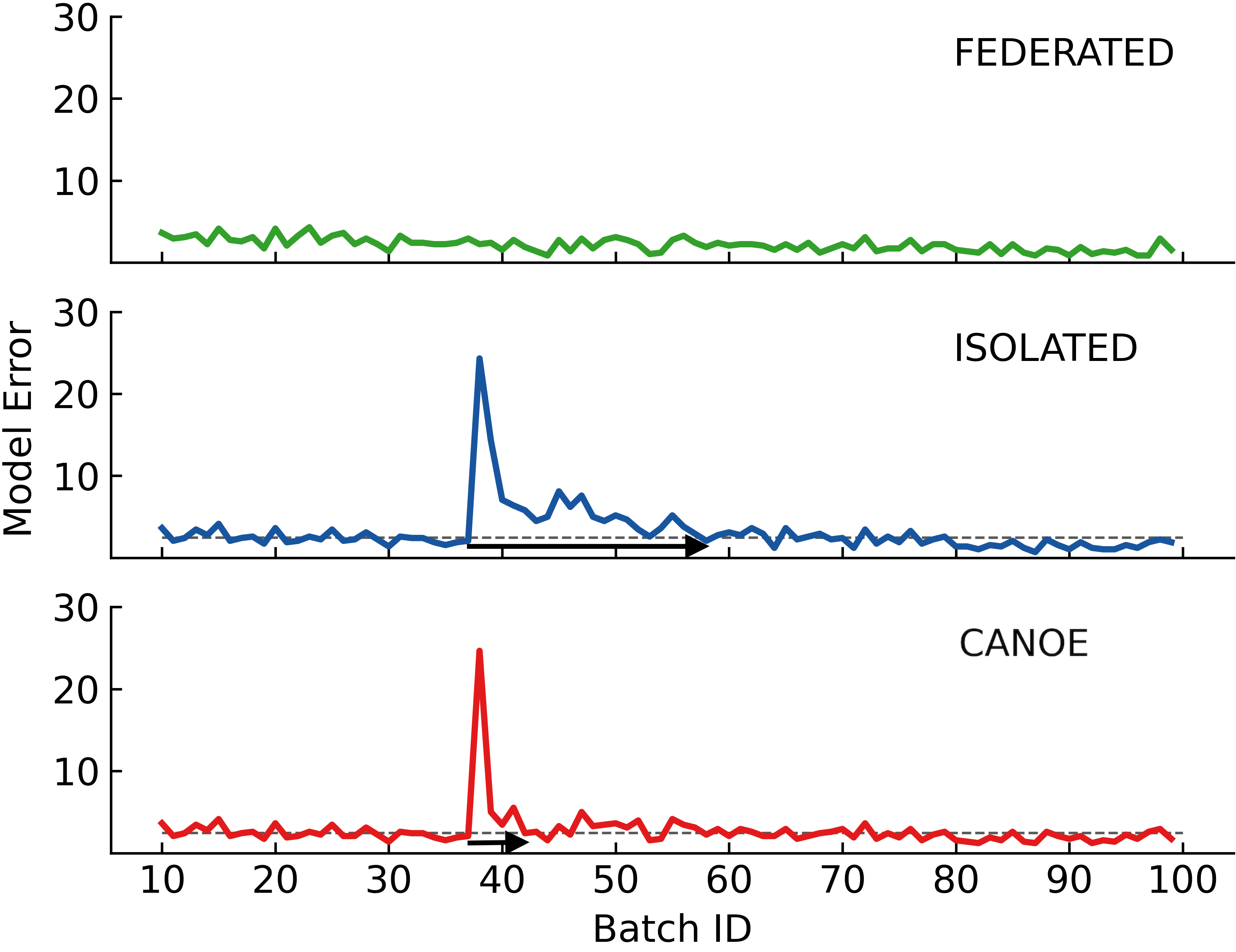}
    \caption{DenseNet}
    \label{fig:densenet_overall_intro}
 \end{subfigure}
 \begin{subfigure}[b]{0.33\linewidth}
    \includegraphics[width=\linewidth]{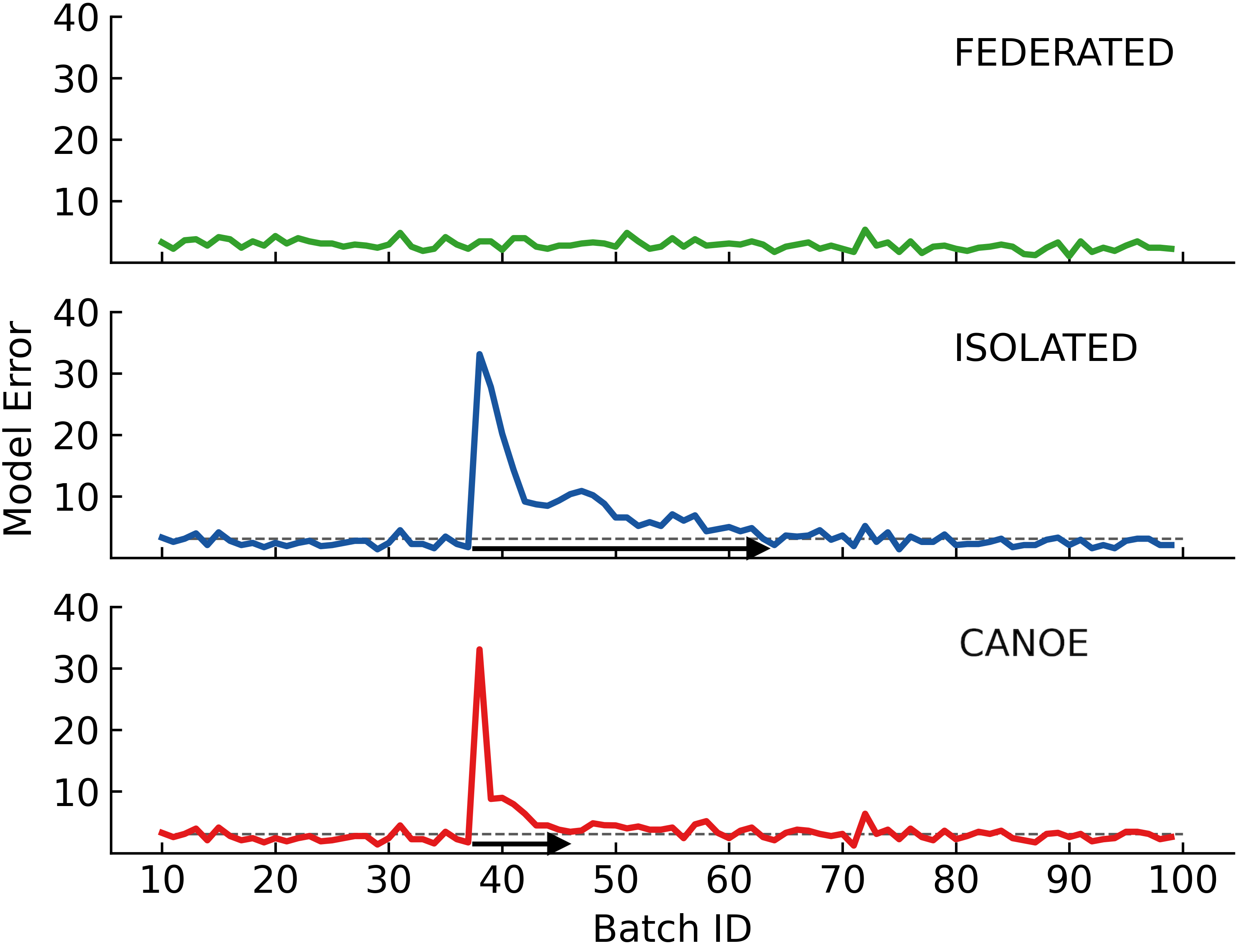}
    \caption{ResNet}
    \label{fig:resnet_overall_intro}
 \end{subfigure}

\caption{Performance comparison for introduction workload distribution (lower error rate is better) showcasing overall model error change. Misclassification of introductory class degrades the model performance. \sys is able to quickly adapt to the change in distribution (given by the horizontal arrow). The horizontal dotted line (in black) defines the lower bound obtained through an offline training.}
\label{fig:overall_perf_intro}
\end{figure*}

 \begin{figure*}[t!]
\centering
 \begin{subfigure}[b]{0.33\linewidth}
    \includegraphics[width=\linewidth]{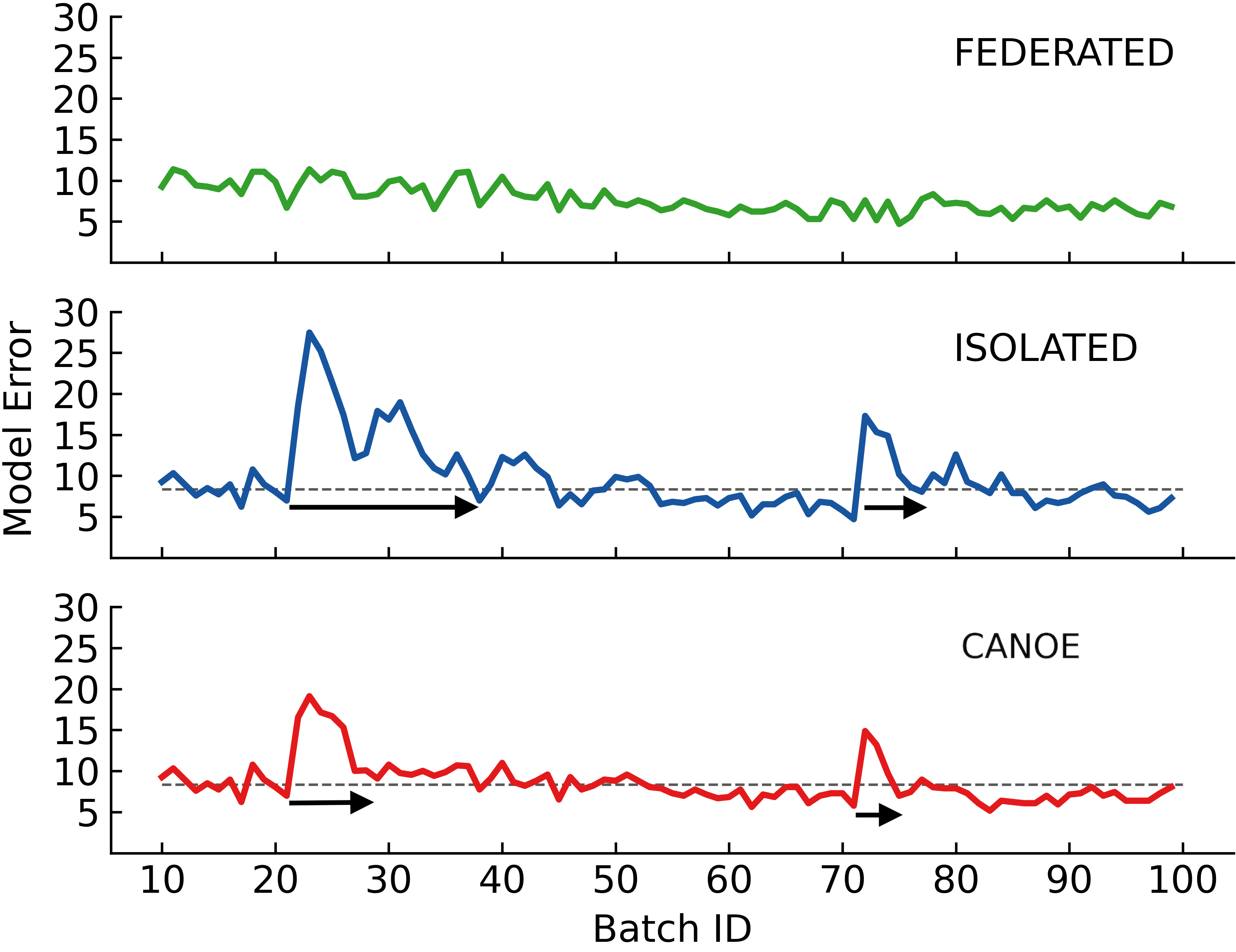}
    \caption{MobileNet}
    \label{fig:mobilenet_overall_fluc}
 \end{subfigure}
 \begin{subfigure}[b]{0.33\linewidth}
    \includegraphics[width=\linewidth]{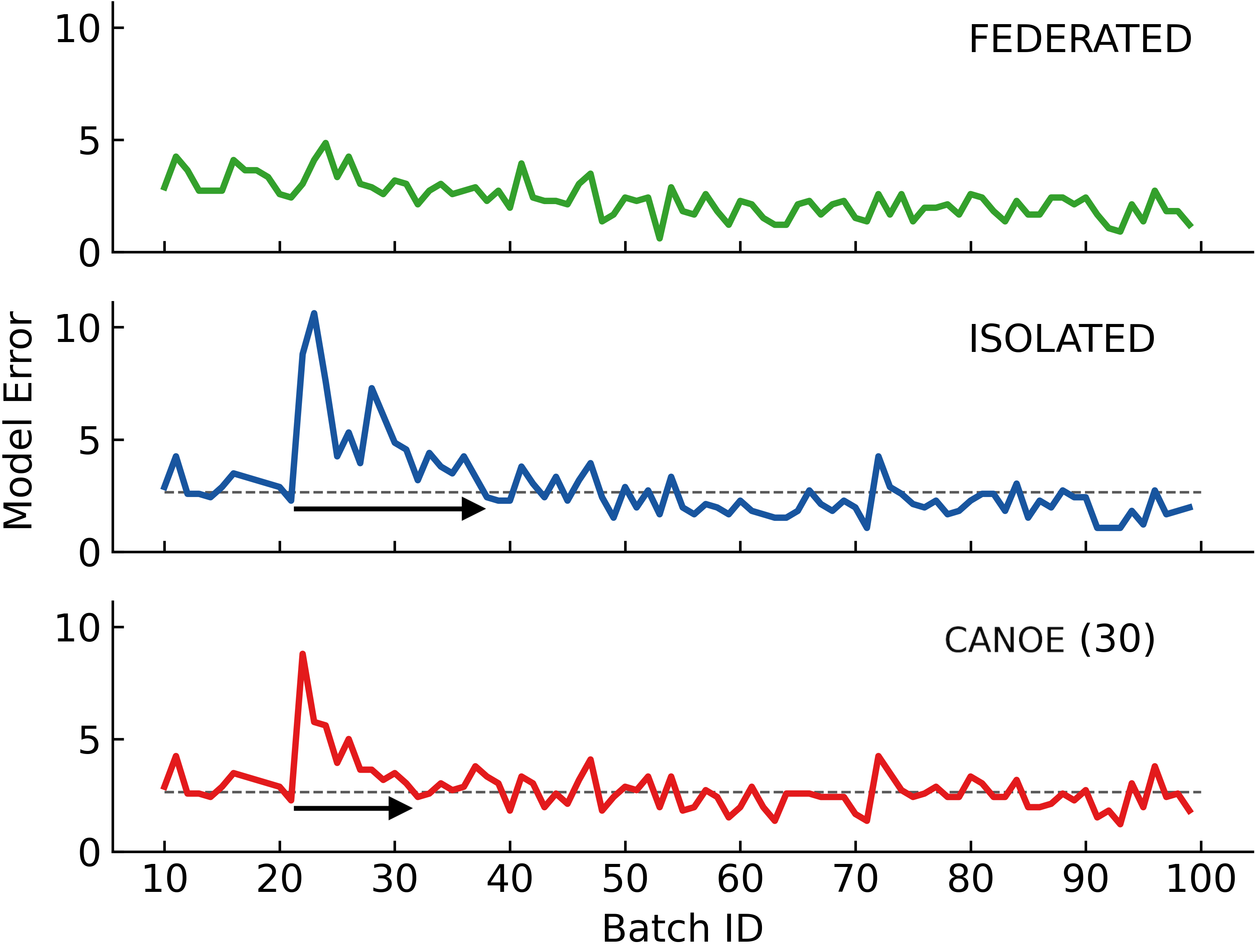}
    \caption{DenseNet}
    \label{fig:densenet_overall_fluc}
 \end{subfigure}
 \begin{subfigure}[b]{0.33\linewidth}
    \includegraphics[width=\linewidth]{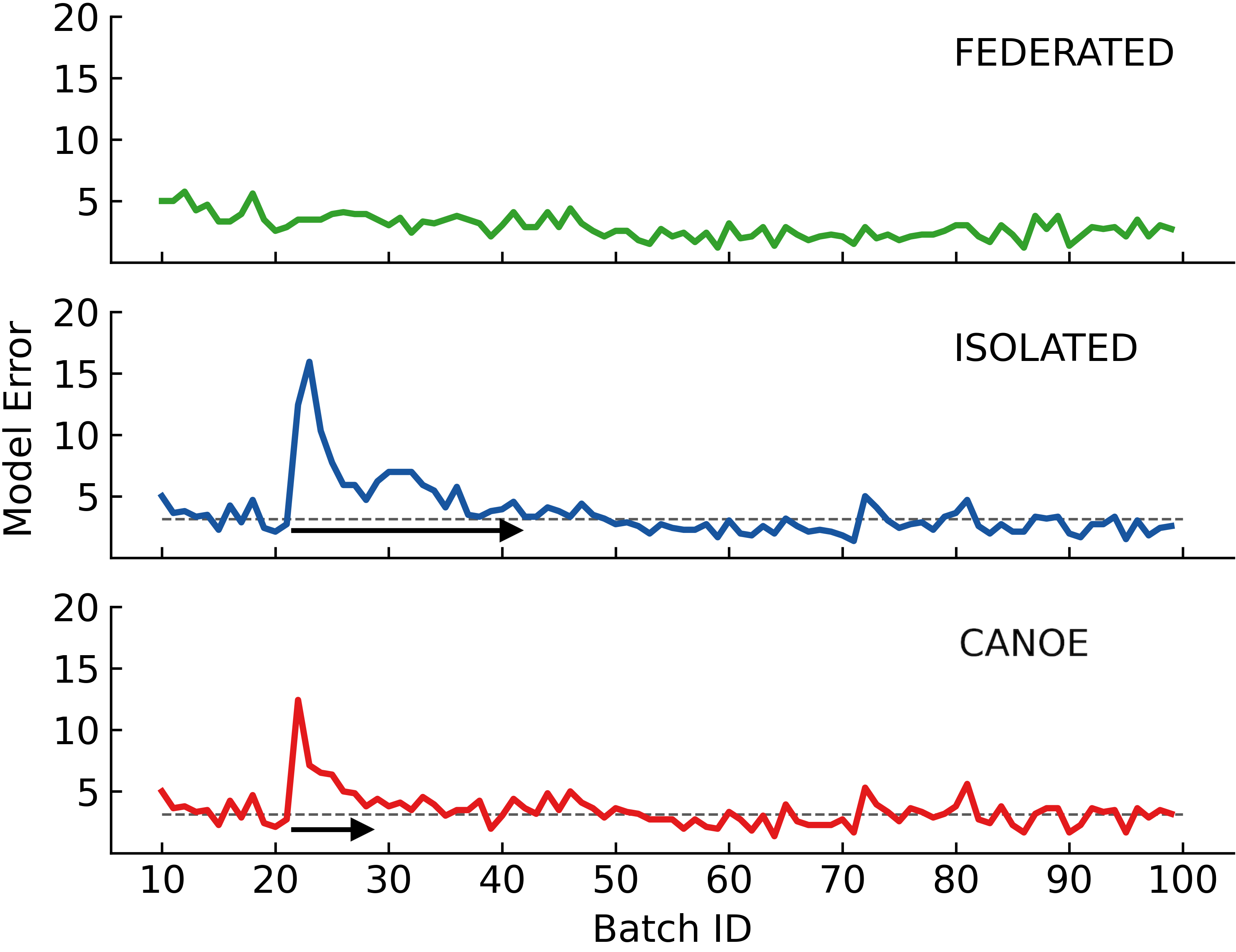}
    \caption{ResNet}
    \label{fig:resnet_overall_fluc}
 \end{subfigure}

\caption{Performance comparison for fluctuation workload. Similar to Figure~\ref{fig:overall_perf_fluc} \sys is able to adapt to the changes in the distribution and behaves close to a federated learning system.}
\label{fig:overall_perf_fluc}
\end{figure*}

\section{Evaluation}
\label{section:evaluation}
The evaluation of \sys explores the following questions:
\begin{tightitemize}
\item[1.] 
 How effective is the knowledge transfer mechanism enabled by \sys in allowing neural network models to adapt to change, and what impact does it have on improving the overall model performance and curtailing data transfer costs? (\S\ref{eval:benefits})
\item[2.] 
  What are the overheads introduced by the parameter selection in \sys on the helper nodes, and by multi-model learning on the target node? 
  (\S\ref{eval:overheads})
\item[3.] 
  What is the impact of the different system-level parameters in \sys
  on the overall benefits?
  (\S\ref{eval:tuning})
\end{tightitemize}

We used multiple workloads, models, and frameworks, as described below,  
  to ensure the generality of the observations. 


  \subsection{Experimental Methodology}
  \label{eval:methodology}
  
  \noindent{\bf Experimental Setup. } We evaluate \sys on a cluster of
  5 nodes on the Chameleon Cloud~\cite{chameleon_cloud} experimental
  platform. All nodes are Intel(R) Xeon(R) 125GiB RAM with 48 cores
  running 2 threads per core at 2.30GHz. 
  The experiments are performed with two versions of \sys, integrated with two existing machine learning frameworks: PyTorch v1.7.1 and TensorFlow v2.3.1 (\S\ref{eval:frameworks}).

  \noindent{\bf Applications, Datasets, Workloads. } We use two application use cases in the evaluation: image classification and intrusion detection (\S\ref{eval:usecase_networks}).

  For the image classification use case the results are obtained with the EMNIST dataset~\cite{emnist}, consisting of 280k data samples. We made similar observations with the smaller MNIST dataset~\cite{mnist}.
These datasets are derived from the NIST Special Database 19~\cite{nist_special_db} where EMNIST follows the same conversion paradigm used to create the MNIST dataset. The outcome is a set of datasets that constitute a more challenging classification tasks involving letters and digits, and that shares the same image structure and parameters as the original MNIST task, allowing for direct compatibility with all existing classifiers and systems. 

For the intrusion detection use case, we used the NSL-KDD
dataset~\cite{nsl-kdd} which consists of benign network requests mixed
with various malicious attacks consisting of 126k records. It is
derived from the KDD'99 dataset~\cite{kdd} where redundant and
duplicate records were removed to avoid bias in the learning
process. 


During the experiments, each edge node processes request batches over
time, with each batch consisting of a  varied number of data points
corresponding to different classes in the dataset. 
To create a change in class priors that
triggers the need for collaboration and knowledge transfer, 
we generate workloads with several synthetic request patterns,
corresponding to {\em introduction} of a new class at an edge node,
{\em fluctuation}, where a new class appears and disappears,
with overlapping and non-overlapping classes, as described in \S\ref{subsection:applying_significant_param}. 
The use of synthetic workload patterns is due to the limited
availability of real infrastructure and data with both temporal and
spatial metadata on the request distribution. The concrete patterns we
use  
are based on information about distributions used in other research~\cite{cartel,att_edge_data,traffic_dynamics_cellular_network,cell_network_traffic}. 
The extensive nature of the experiments successfully demonstrates the
benefits achieved from the mechanisms provided in \sys.

\noindent{\bf Models.}
\label{subsection:models}
For the image classification use case,
most of the
presented results are based on 
PyTorch and
the MobileNet~\cite{mobilenet}, DenseNet~\cite{densenet} and
ResNet~\cite{resnet} 
CNN models. The models
are fine tuned using 2k data points from the EMNIST dataset using Adam
optimizer and learning rate 0.00005, 
and trained for 20
epochs with batch size of 30. Their sizes and parameter counts are
summarized in Table~\ref{table:per_request_data_transfer}.
%
We also used the MobileNet, DenseNet and InceptionNet~\cite{inceptionnet} models provided by TensorFlow v2.3.1, to confirm the effectiveness of \sys across different machine learning frameworks (\S\ref{eval:frameworks}).

The intrusion detection use case uses a four-layer DNN model with
209k parameters, 
as described in~\cite{dnn_network_analysis}. For model initialization we used the SGD optimization algorithm with a learning rate is 0.000005 and 25k data points. The model is trained for 20 epochs with batch size of 30.

\noindent{\bf Metrics.}
When using local, tailored models, as is the case in isolated or
collaborative learning, a change in the workload pattern at a node via
the appearance of previously unseen classes, results in a temporary
drop in the predictive performance of the model, observed as an
increase in its error rate. 
The time taken for the model to reach back the acceptable performance levels is defined as the model's {\em adaptability to change}.
Models using 
FL are usually impervious to such changes, because the frequent model update from the aggregator node 
in the FL framework continuously distributes the knowledge obtained at
one node to all the other nodes in the system. 
In isolated learning, no knowledge is received from any other node in the system, instead the model performance ultimately improves as a result of the online learning (retraining) on the data from the incoming request batches. 
The goal of \sys is to improve the model's adaptability by minimizing the time required to adapt to change compared to learning in isolation, while also reducing the data transfer requirements compared to federated learning. 

We evaluate \sys by measuring the improvements in adaptability to
change it offers compared to isolated learning. Given the synthetic
nature of the workloads, we measure time in terms of the number of
batches that need to be processed for the model performance to reach
the same error rate as what is possible with FL. We also measure the
improvements that \sys provides in terms of per-node and aggregate
data transfer costs (in bytes). Finally we measure the runtime
overheads of \sys in terms of additional compute (using processing time as a measure) and memory resources. 

\subsection{Benefits of \sys}
\label{eval:benefits}

\begin{table*}[t!]
\small
\centering
\caption{Raw data transferred (MB)  and gains per {\em update} per
  node in FL vs.~\sys. In Federated Learning data is transferred to send and received model update while in \sys data is transferred to share metadata information and during knowledge transfer. } 
\begin{tabular}{ p{2.5cm}|p{1.8cm}|p{1.5cm}|p{1.5cm}|p{1.5cm}|p{1.5cm} |p{1.8cm}|p{2cm}} 
	\multirow{2}{*}{\textbf{CNN Model}}& \multirow{2}{*}{\parbox{4cm}{\textbf{Million \\ Parameters}}} & \multicolumn{2}{c}{\textbf{Federated Learning}} & \multicolumn{2}{|c}{\textbf{\sys}} & \multicolumn{2}{|c}{\textbf{Data Transfer Gains}}\\
		\cline{3-8}
		& & \textbf{Out} & \textbf{In} & \textbf{KT (In)} & \textbf{MdS (Out)} & \textbf{No Drift ($\times$)} & \textbf{With Drift ($\times$)}\\
 \hline
 MobileNet & 3.54 & 6.38 & 7.36 & 4.57 & 0.00058 & 23684 & 3\\ 
 DenseNet & 8.06 & 16.18 & 18.35 & 8.80 & 0.00058 & 59507 & 4\\  
 ResNet & 11.69 & 12.19 & 13.75 & 8.55 & 0.00058 & 44713 & 3\\  

\end{tabular} 
\label{table:per_request_data_transfer}
\end{table*}

\begin{table}[t!]
\small
\centering
\caption{Ratio of total data transferred (in and out) of a node in Federated Learning versus \sys for a 5 node setup for an entire experiment consisting of 100 request batch.}
\begin{center}
	\begin{tabular}{p{1.5cm}|p{1.8cm}|p{2.1cm}|p{1.5cm}}
    	\textbf{Model} &  \textbf{Workload} & \textbf{Aggregator($\times$)} & \textbf{Member($\times$)} \\
    	\hline
		 \multirow{2}{*}{MobileNet} & Introduction & 1062 & 267\\ 
    	  & Fluctuation & 644 & 161\\     	  
    	 \multirow{2}{*}{DenseNet} & Introduction & 1897 & 350 \\ 
    	  & Fluctuation & 1396 & 346 \\ 
    	  
    	  \multirow{2}{*}{ResNet} & Introduction & 1078 & 271\\ 
    	   & Fluctuation &	1069 & 	268 \\  
	\end{tabular}
\end{center}
\label{table:data_transfer_summary}
\end{table}

\noindent{\bf Adaptability to Change. }
Figures~\ref{fig:overall_perf_intro} and~\ref{fig:overall_perf_fluc} illustrate the adaptability enabled by \sys for the {\em introduction} and {\em fluctuation} workloads where the system uses $50\%$, $20\%$ and $30\%$ of the significant parameters for MobileNet, DenseNet and ResNet, respectively.
The parameter $Z$ that controls the percentage of knowledge transferred for each of the three models is experimentally 
chosen to that it 
minimizes data transfer while keeping the model adaptability within 10\% of the best case.
For the given testbed configurations, in these experiments the helper model is created after within 4 request batches after the drift detection.
We observe that with \sys, 
the models  converge to the performance level equivalent to using a global model up to $2$ to $3.5\times$ faster as compared to an isolated system. 
%
%
For the {\em fluctuation} workload and MobileNet, in Figure~\ref{fig:mobilenet_overall_fluc} we observe a second spike in the model performance as the class reappears in the later part, while 
for DenseNet and ResNet 
we do not observe any change for the same workload. This is because the other two models are larger and are able to retain more information even when the class was not present for few request batches. 


\noindent{\bf Reduction in data transfer costs. } The collaborative learning technique used by \sys works on the curve of finding a tradeoff at spending few cycles to adapt to the change in the workload versus the data transfer cost associated in regularly creating and updating a global model, as done in FL.
We summarize these costs in Table~\ref{table:per_request_data_transfer}. 
In FL, for each update period, for a member node the data transfer cost consists of the delta in the local model since the last batch ({\em Out}) and the received 
aggregated model update ({\em In}). 
For \sys, the per-update data transfer cost at each node consists of the metadata 
sent to the MdS server ({\em MdS(Out)}).  When data drift is detected on a node, that node also incurs the knowledge transfer cost ({\em KT}).
For neural networks, the change in a local model ($Out$) or aggregated model updates ($In$) are significantly larger when compared to the metadata (few kBs) used by \sys, and this leads to drastic reductions in the data transfer costs, compared to FL, 
up to $10^4\times$, as seen from the table.
Moreover, since only a fraction of model parameters are transferred as part of knowledge transfer in \sys, this too could result in data saving of up to $4\times$ when compared to a single model update request in FL.

In Table~\ref{table:data_transfer_summary},  we compare the total data transfer for the same experiment with 5 edge nodes with FL and \sys as used in Figures~\ref{fig:overall_perf_intro} and~\ref{fig:overall_perf_fluc}. 
As the collaborative approach requires model data to be exchanged only when there is a drift in the model performance at a target node, while in FL all nodes frequently receive model update across the system the outcome is an overall data transfer reduction of 
2-3 orders of magnitude, depending on the the workload and model.

\subsection{\bf Cost of \sys Mechanisms}
\label{eval:overheads}

\noindent{\bf Extracting significant parameters. }
Using \sys to perform knowledge transfer introduces runtime overheads associated with the execution of the necessary mechanisms. We present these costs in the context of the same experiments as above using the {\em introductory} workload.

The {\em continuous} method for parameter selection
stores the sensitivity value of all parameters for all of the classes, and its overheads depend on the number of layers and parameters of the model. 
For our dataset and given models, this introduces additional memory overhead of 
 $140MB$, $320MB$, $470MB$ for MobileNet, DenseNet and ResNet, respectively.
 The {\em on-demand} approach creates a local datastore 
 which is only used when a request from a target node is received. For our dataset, with an average of 600 data points in a request batch, storing only the most recent request batch was sufficient and required an addition of $6.8MB$ of memory.
We performed the same experiments for different batch sizes, starting from as low as 50 data points, and, across workloads, observed similar gains in the overall benefits of \sys, with expected reduction in the runtime overheads.

In terms of the compute overhead, the 
{\em on demand} approach 
adds an overhead in the response time of $0.9$ to $1.35\times$ the time required to process a request batch. 
The {\em continuous} approach by pre-calculating these values avoids such delays in the critical path of response, however, adds an overhead of up to $0.7\times$ the time required to process a request batch, for every request, 
as shown in Table~\ref{table:mechasim_overhead}. 

\begin{table}[t!]
\small
\centering
\caption{Overhead in performing mechanisms described in Section~\ref{section:mechanisms} normalized with the time required to process a request batch.}
\begin{tabular}{ p{1.2cm}|p{2.2cm}|p{2.2cm}|p{1.4cm} } 
  \textbf{Model} & \textbf{Continuous ($\times$) per batch} & \textbf{On-Demand ($\times$) per request} & \textbf{Helper Model ($\times$)} \\ 
 \hline
 MobileNet & 0.70 & 0.98 & 0.08 \\ 
\hline
 DenseNet & 0.65 & 1.35 & 0.11 \\  
 \hline
 ResNet & 0.70 & 0.95 & 0.14 \\  
 \hline
\end{tabular} 
\label{table:mechasim_overhead}
\end{table}

\noindent\textbf{Helper model creation.} The time taken to create the model depends on the size of the portion of model transferred from the helper node, as shown in Table~\ref{table:per_request_data_transfer}, and the time taken to create a model at the target node. Upon receiving the significant parameters creating a helper model takes only up to $0.14\times$ the time required to process a request batch.

\begin{figure}[t!]
    \centering
    \includegraphics[width=\columnwidth]{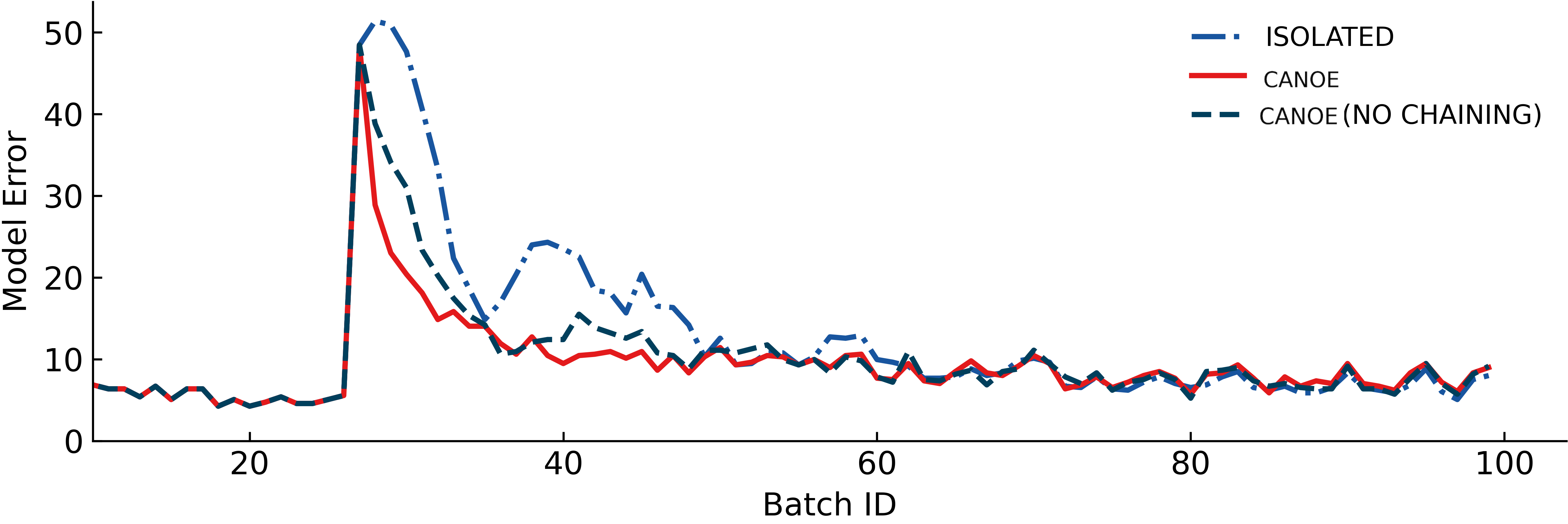}
    \caption{Chain of helper models improves the predictive performance and helps the system adapt to the new changes faster.}
    \label{fig:model_chain}
\end{figure}

\noindent\textbf{Management of helper models.}
A helper model in \sys uses similar memory as the resident node model however since the model is frozen, no additional resources are required after its creation. Using multiple helper models in a chain adds to the memory overheads. 
Thus, if the knowledge about the introductory classes at the target node is dispersed among multiple nodes, in such a case \sys exposes a tradeoff
of using more memory for a short duration of time by deploying chains of models vs.~spending more time in adapting to the changes. Figure~\ref{fig:model_chain} shows the collaborative performance of using help from only one of the nodes (no chaining) compared to creating chain of models from multiple nodes. 
The ability to combine multiple knowledge sources  in \sys provides the lower bound on adaptability to the new changes in the system.

\subsection{Sensitivity Analysis \& Opportunities}
\label{eval:tuning}

The benefits of \sys 
are coupled with the quantity and quality of knowledge transferred. 
We analyze this by considering the performance of the models and workloads evaluated in \S\ref{eval:benefits}, and varying the amount of knowledge transferred and the quality of the helper node used for knowledge transfer. 

\begin{figure}[t!]
\centering
 \begin{subfigure}[b]{0.49\linewidth}
    \includegraphics[width=\linewidth]{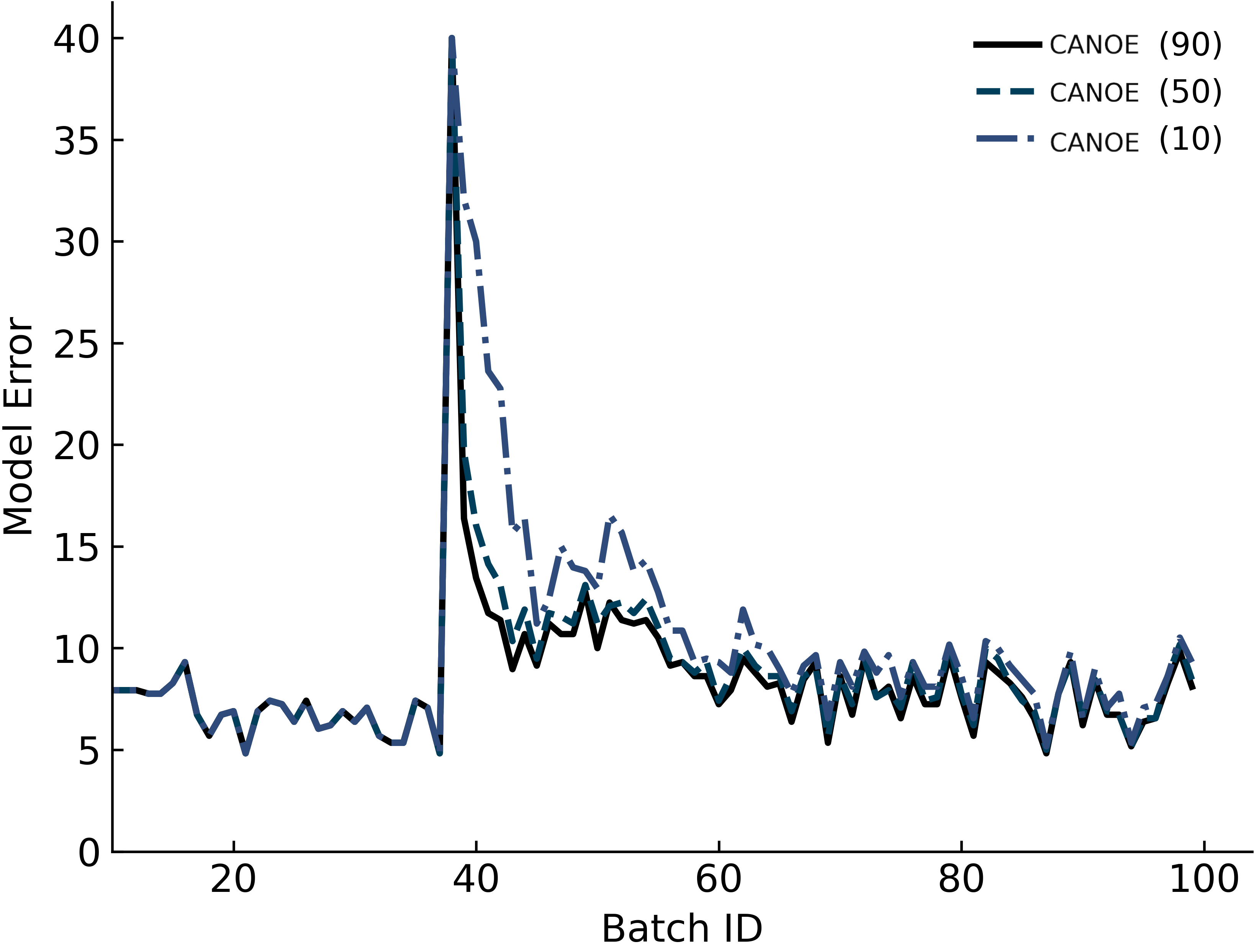}
    \caption{Significant parameter selection}
    \label{fig:sp_knob}
 \end{subfigure}
 \begin{subfigure}[b]{0.49\linewidth}

	\includegraphics[width=\linewidth]{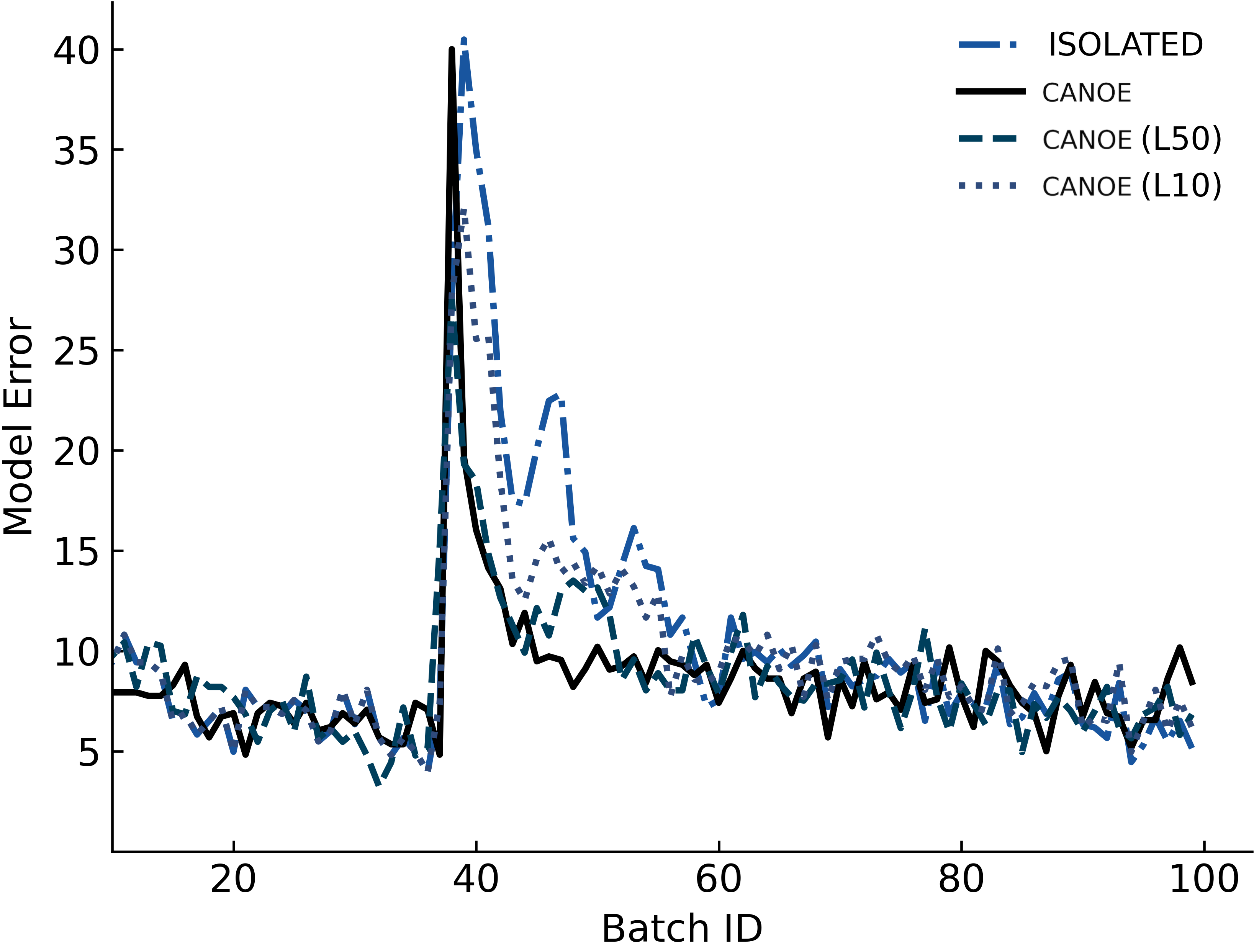}
    \caption{Helper node quality}
    \label{fig:varying_ln}
 \end{subfigure}
\caption{Comparison of impact of percent of significant parameter $Z$ (left) and impact of quality of helper node model (right).}
\label{fig:significant_parameter_and_data_transfer_knob}
\end{figure}

\vspace{0.6ex}
\noindent\textbf{Impact of quantity of knowledge transferred.}
\sys
controls the portions of the significant parameters that are
transferred via the parameter $Z$.
We evaluate the impact of different values of $Z$ on the model's 
adaptability and data transfer costs.   
Figure~\ref{fig:sp_knob} shows the results of evaluation for {\em introductory} workload pattern for MobileNet; the results with the remaining models had a lower sensitivity to $Z$ than what we present for MobileNet. The results illustrate two important aspects of \sys. An effective helper model can be created using only a portion of the relevant parameters. The percent of significant parameters received from the helper node have a direct impact on improving the target node's overall predictive performance and reduction the data transfer cost. 
The higher value of $Z$ (90) may provide improvements in error rate,
when compared to $Z$ (10), specifically for this setup 25\% reduction
in average model error rate while using the helper model. However, it
would result in $5.53\times$ more data transfer. 



We note that \sys introduces new ability to control a tradeoff among learning performance and data transfer costs, by using the amount of knowledge that should be transferred across nodes as a configuration knob. This can be leveraged by future learning management policies to adapt the effectiveness of the process based on the model used and the deployment scenario.
  
\noindent\textbf{Impact of quality of helper node.} 
As discussed in \S\ref{section:background_motivation} a helper node
can be identified from the model and node metadata, however, its
choice can have an impact on the collaborative learning process. To
understand this, in Figure~\ref{fig:varying_ln}, we compare the
execution of \sys and Z(50) with MobileNet and the same helper node as
used in the previous experiment, to that of two other scenarios where
the helper nodes have observed only 50\% (L50) and 10\% (L10) of the
data observed in the baseline. 
As expected, we observe that the quality of the helper node does
impact the realized benefits from \sys. However, we see that using
even a weaker helper node (L10) can still be useful, compared to just
learning in isolation. Note that, simply increasing $Z$ for a weak
helper node would not be helpful, as the quality of the model parameters is not good for the required knowledge. 

\subsection{Extensibility to Different ML Frameworks}
\label{eval:frameworks}
To demonstrate the generality of the system, \sys is implemented for PyTorch (PT) and TensorFlow (TF).
With TF, \sys creates a pool of helper models from the initial start of the learning process, which are later updated using the significant parameters received from the helper node. Parameter sensitivity values are computed using the {\em continuous} technique. 
Due to these limitations, and for brevity, 
the earlier section presents the results from
PT only. Based on experiments with 
the {\em introduction} and {\em fluctuation} workloads used with the PT 
evaluation, and the MobileNet, DenseNet and InceptionNet CNNs, \sys adapts to change up to 3$\times$ faster than isolated learning, while using only $60\%$, $30\%$ and $30\%$ of the model respectively. This results in overall data transfer reduction of $700$ to $1200\times$ compared to using FL.

\subsection{Use Case -- Intrusion Detection}
\label{eval:usecase_networks}

\begin{figure}[t!]
\centering
 \begin{subfigure}[b]{0.49\linewidth}
    \includegraphics[width=\linewidth]{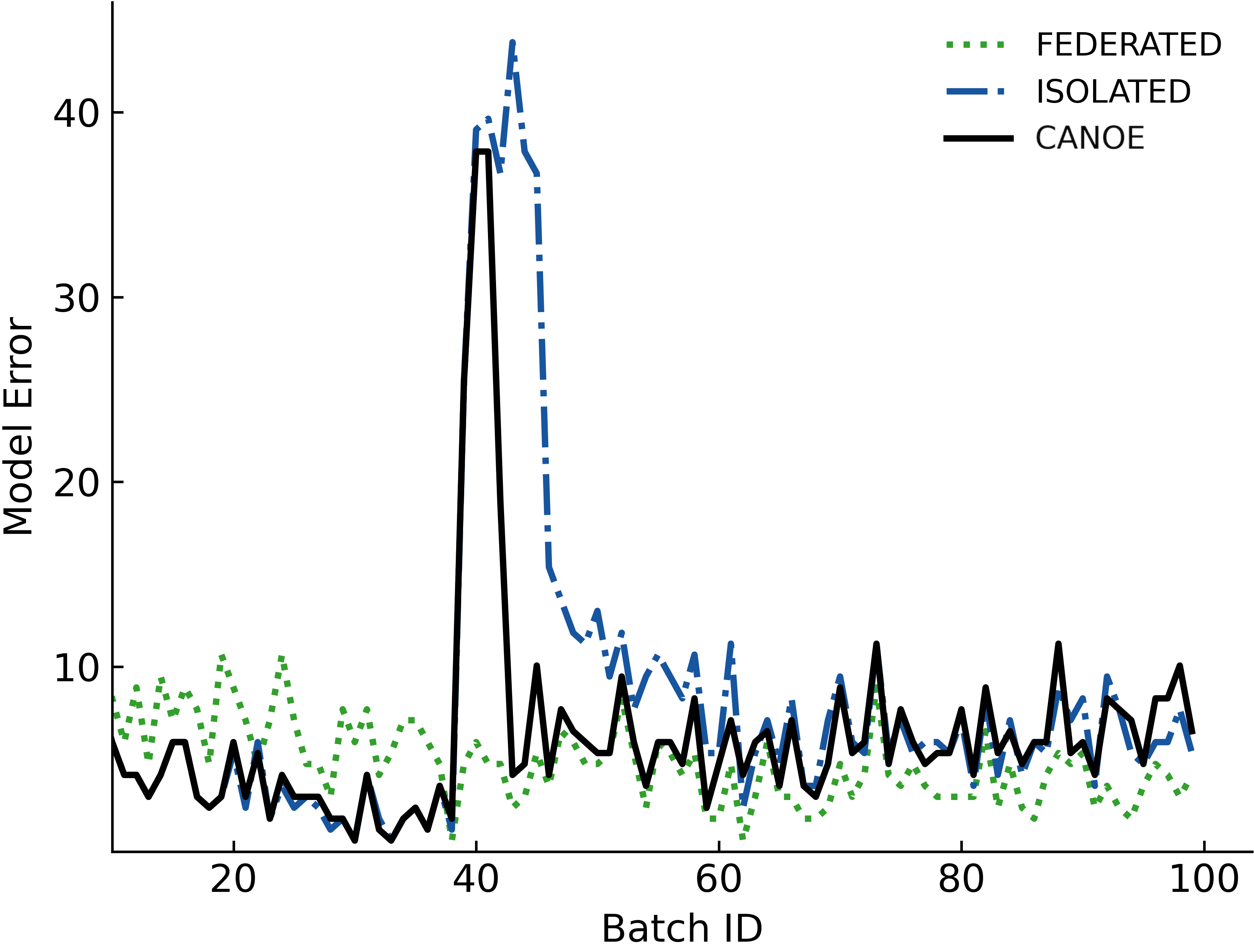}
    \caption{Overall Error}
    \label{fig:usecase_networks_overall_error}
 \end{subfigure}
 \begin{subfigure}[b]{0.49\linewidth}
    \includegraphics[width=\linewidth]{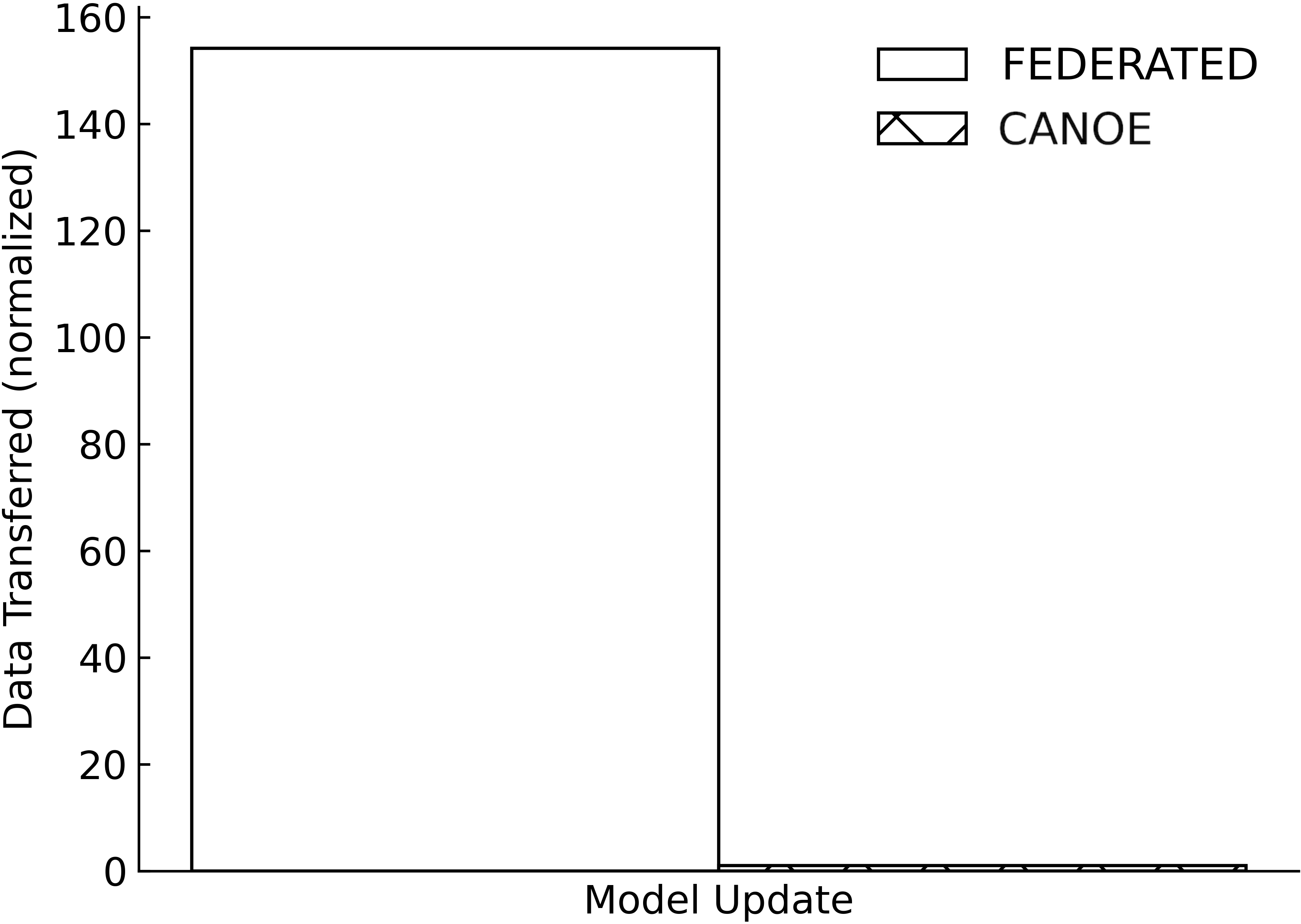}
    \caption{Data Transfer}
    \label{fig:usecase_networks_data_transfer}
 \end{subfigure}
\caption{Performance and total data transfer comparison for network attack dataset using ORF to classify begin request against DDoS and port scan attack requests.
Comparison of model performance and data transfer overhead for network attack dataset using {\em introductory} workload.}
\label{fig:usecase_networks}
\end{figure}

  Intrusion Detection System monitor and analyze the network to detect anomalies and attacks. This use case is based on the NSL-KDD dataset~\cite{nsl-kdd} which consists of four types of attacks: Denial of Service Attack (DoS), User to Root Attack (U2R), Remote to Local Attack (R2L), and Probing Attack (Probe). The testbed consists of 5 edge nodes, where initially the target node
  observes only the R2L and Probe attacks, while the other nodes experiences all of the  attacks. After some time the target node starts to experience the DoS attack. We use a four layer Deep Neural Network model as described in \S\ref{subsection:models} with \textit{introduction} workload consisting of 100 request batches with $\sim$400 data points in each batch and $Z$(40).
  

\noindent\textbf{Results.} Figure~\ref{fig:usecase_networks_overall_error} illustrates the model performance for FL, \sys and isolated. 
FL allows the target node to classify the attack immediately,
since the attack was observed by other nodes earlier, and is captured in the global model updates. 
In isolated learning mode, the target node experiences both a large spike in its model error rate, and takes substantial time (equivalent of 20 batches) to retrain its model to the original performance (error rate). 
\sys bridges the gap between the above two system.
The collaborative process helps the target node adapt to the new attack $1.75\times$ faster than isolated system and to provide similar average model error rate compared to federated learning system with $600\times$ less data transferred 
over the wire.

\section{Related Work}
\label{section:related_work}

\noindent{\bf Distributed Machine Learning.}
\noindent Since its introduction, Federated Learning has emerged as a go-to approach to distributed machine learning~\cite{fl}. Follow-on work~\cite{konevcny2015federated, arivazhagan2019federated, bonawitz2017practical} has introduced a number of optimizations, including recent solutions which speed up the convergence and improve the accuracy through guided member node selection based on developer specified criteria, as done in Oort~\cite{oort}. 
Different from \sys, FL approaches have a goal of learning a generic ML model that will be used by all nodes in the distributed system, which justifies the data transfer costs.


%

To address the data transfer overheads in geo-distributed systems, Gaia \cite{hsieh2017gaia} distinguishes among communication over local (LAN) vs.~wide area networks (WAN).
Nodes within LAN 
use a strong consistency model as used in FL, but only significant model updates 
are shared among WANs to save communication and speed up training. The goal of identifying a significant model update in Gaia is orthogonal to that of identifying significant model parameters, as the former applies to the model overall, whereas the technique used in \sys is concerned with identifying a portion of the model that most contributes to the model quality with respect to a specific class. As with FL, the goal of Gaia is to learn a global model, shared across the distributed system.

In contrast, collaborative distributed learning approaches described in \S\ref{section:background_motivation}, are developed for environments where a global model is not needed \cite{dlion,colla,cartel,online_distillation}.
\sys contributes to such approaches by proposing a generic system support that makes
them applicable to scenarios based on 
neural network models.
%
%
Other types of distributed machine learning system introduce complementary techniques aimed at model specializations \cite{qin2019task, yu2018distilling, wu2020personalized}, including when targeting edge nodes and IoT, as well as for addressing privacy concerns about data and model sharing \cite{merugu2003privacy, so2019codedprivateml, huang2019dp}.

\noindent{\bf Knowledge Transfer Technique.}
\sys uses significant parameters of an ML model to perform knowledge transfer.
The mechanisms used in \sys are similar to other approaches to determining significant parameters from a model \cite{rafegas2020understanding, yu2018distilling,simonyan2013deep, zhou2018revisiting},  but are simplified so as to provide for fast calculations and enable their online use. Note that \sys is intended to be a general framework that can be used with different techniques for significant parameter selection, with a corresponding impact on the benefits or costs of learning.

Boosting \cite{freund1999short} is used as a technique to combine the output from more than one models, to adapt to change when drift happens~\cite{dai2007boosting, oza2005online}, 
or to speed up the model responsiveness~\cite{clipper}. 
Instead of training and storing multiple models 
at each edge, \sys 
uses boosting in conjunction with dynamically created helper models, so as to improve their effectiveness. 


\section{Conclusion}
\label{section:conclusion}

\noindent 
We introduce \sys, a system that enables
collaborative learning for deep neural networks. To achieve this goal,
\sys contributes new support for knowledge transfer across neural
networks, which makes it possible to dynamically extract helpful
knowledge from one node -- in the form of significant model parameters
-- and to apply it to another -- using dynamically created helper
models and multi-model boosting.
\sys is prototyped and evaluated with several neural network models for the
PyTorch and TensorFlow ML frameworks. The experimental results
demonstrate that \sys can improve the model's adaptiveness when a drift occurs, while maintaining low communication cost. Furthermore, it provides
new interfaces to exercise the tradeoffs among accuracy and overheads
in the learning process. This can expose flexibility in how
learning can be tuned to different use case requirements
that can be exploited in future resource
management policies.

\bibliographystyle{ACM-Reference-Format}
\bibliography{main}

\end{document}